\documentclass{article}

    \PassOptionsToPackage{numbers, compress}{natbib}

 \usepackage[eandd, preprint]{neurips_2026}

\usepackage[utf8]{inputenc} 
\usepackage[T1]{fontenc}    
\usepackage{hyperref}       
\usepackage{url}            
\usepackage{booktabs}       
\usepackage{amsfonts}       
\usepackage{nicefrac}       
\usepackage{microtype}      
\usepackage{xcolor}         

\usepackage{graphicx} 
\usepackage{caption}
\usepackage{subcaption}
\usepackage{setspace}        
\usepackage{microtype}       
\usepackage{xcolor}          
\usepackage[utf8]{inputenc}
\usepackage{booktabs}
\usepackage{multirow}
\usepackage{xcolor}
\usepackage{array}
\usepackage{colortbl}
\usepackage{graphicx}
\usepackage{multirow}
\usepackage{amssymb}
\usepackage{pifont}
\usepackage{makecell}
\usepackage{subcaption}
\usepackage{amsmath}
\usepackage{amssymb}
\usepackage{mathtools}
\usepackage{amsthm}
\usepackage[utf8]{inputenc}  
\usepackage[most]{tcolorbox}
\usepackage{xcolor}
\newtcolorbox{promptbox}{
  enhanced,
  breakable,
  colback=gray!10,
  colframe=black,
  boxrule=0.8pt,
  arc=2pt,
  left=8pt,
  right=8pt,
  top=8pt,
  bottom=8pt,
  fontupper=\ttfamily\scriptsize\raggedright,
  width=\linewidth,
}
\usepackage{listings}
\usepackage{xcolor}
\usepackage[most]{tcolorbox}

\lstdefinestyle{codestyle}{
  basicstyle=\ttfamily\small,
  backgroundcolor=\color{gray!10},
  frame=single,
  rulecolor=\color{black},
  framerule=0.8pt,
  xleftmargin=8pt,
  xrightmargin=8pt,
  aboveskip=8pt,
  belowskip=8pt,
  breaklines=true,
  breakatwhitespace=false,
  columns=fullflexible,
}

\hypersetup{
    colorlinks=true,
    citecolor=blue,
    linkcolor=blue,
}

\newcommand{\pmark}{\includegraphics[height=0.8em]{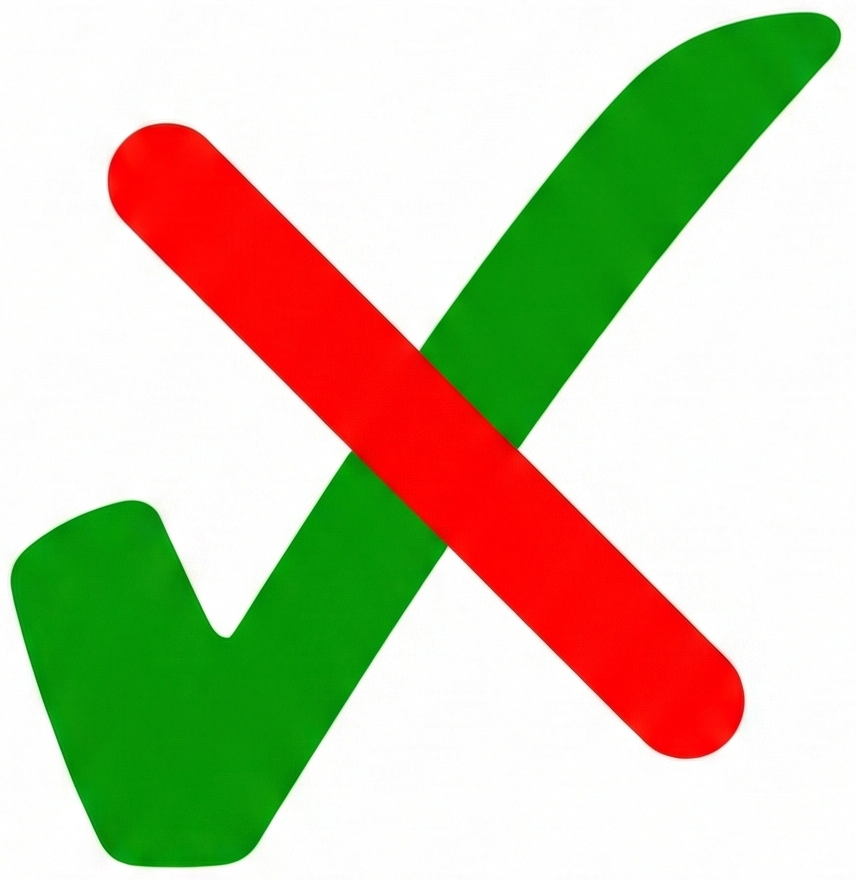}}
\definecolor{vividPurple}{RGB}{128, 0, 200}
\definecolor{rosePink}{RGB}{255, 105, 135}
\definecolor{myPink}{RGB}{220, 80, 120}

\title{Pix2Fact: When Vision Is Not Enough — Benchmarking Fine-Grained VQA with Web Verification on High-Resolution Real-World Scenes}

\author{
Yifan Jiang\textsuperscript{1,2}, 
Cong Zhang\textsuperscript{1,3}\thanks{Corresponding Author: cong.zhang92@gmail.com},\ \  
Bofei
Zhang\textsuperscript{1,4}, 
Yifan Yang\textsuperscript{3},\\
\textbf{Qiaofeng~Zheng\textsuperscript{5}, Bingzhang~Wang\textsuperscript{6}, and
Yew-Soon~Ong\textsuperscript{3}}\\
\textsuperscript{1}GADE Union (Global AI Data Experts Union),
\textsuperscript{2}Shanghai Jiao Tong University, China,\\
\textsuperscript{3}Nanyang Technological University, Singapore,
\textsuperscript{4}New York University,\\
\textsuperscript{5}Cambridge University,
\textsuperscript{6}The University of Hong Kong\\
}

\begin{document}

\maketitle

\begin{abstract}
Despite progress on general tasks, vision-language models (VLMs) still struggle with challenges that demand both fine-grained visual grounding and external knowledge, a synergy overlooked by existing benchmarks that evaluate these abilities in isolation. To fill this void, we introduce Pix2Fact, a visual question-answering benchmark designed to assess expert-level visual perception and knowledge search. Pix2Fact comprises 1,000 high-resolution (4K+) images spanning eight scenarios. Its questions and answers are meticulously crafted by PhD-holding annotators from top global universities across diverse disciplines. Each question requires detailed visual grounding and the integration of external knowledge. Evaluating ten state-of-the-art VLMs, including proprietary models such as Gemini-3.1-Pro and GPT-5.4, we find that Pix2Fact poses a formidable challenge: the most advanced model (Gemini-3.1-Pro) achieves only 51.7\% average accuracy, even with access to visual ground truth and search tools. Our analysis attributes this low accuracy to three factors, frequent visual grounding errors even with visual ground truth, shallow search harnessing, and VLM's inability to retrieve long-tail, unstructured local information. This striking gap exposes the limitations of current models in assisting humans with real-world scenarios that demand overwhelming visual comprehension. We believe Pix2Fact will serve as a critical benchmark to drive the next generation of language-vision agents that seamlessly integrate fine-grained perception with robust knowledge search.
\end{abstract}

\section{Introduction}

The evolution of AI models is moving increasingly toward task-specific specialization~\cite{comanici2025gemini,anthropic2024claude,10.1109/MCI.2023.3277769}, not the fragmentation of isolated abilities. For complex domains like visual understanding, solving real problems demands multiple integrated capabilities, such as reasoning and search, within a single unified model. This shift exposes a critical gap in evaluation, as current benchmarks fail to holistically assess how effectively a model combines these diverse skills in practice. 

This challenge is particularly evident in visual language models (VLMs). Accurately grounding small objects in high-resolution scenes remains a critical, last-mile obstacle to realizing their promise in human daily-living applications. Although recent progress in high-resolution visual grounding has been driven by benchmarks focusing on structured domains like graphical user interfaces (GUIs) for tasks such as operating system control~\cite{zhao2025worldgui}, a comprehensive benchmark for open-world, daily-life scenarios is still notably missing~\cite{gao2025omniground,zhao2025rgbt,wu2024v}. This gap not only hinders our ability to systematically assess and improve vision-language agents in unstructured, knowledge-rich environments but also presents a significant barrier to their practical application in assisting humans across real-world scenes.

A quintessential example from daily life is a tour guide identifying a distant mountain and recalling its history upon a tourist's query. For humans, this task, i.e., effortlessly linking subtle visual cues with structured facts via broad knowledge, is almost automatic, often aided by a reference like a guidebook. For modern VLMs, however, it remains extremely difficult. The challenge lies not in visual grounding or knowledge search alone, but in their tight coupling: the model must simultaneously ground the query in small, local visual details and execute a deliberate, multi‑step knowledge search to retrieve and synthesize the correct information. Precisely because real‑world scenarios demand this integration, the combined requirement of fine‑grained visual perception and open‑world knowledge search remains a significant, yet unmeasured, bottleneck for current Vision‑Language Models (VLMs)~\cite{lin2025vision,bai2025multi}. To close this gap, we introduce Pix2Fact, a new visual question‑answering benchmark designed to evaluate the conjunction of fine‑grained visual grounding and deliberate, open‑world knowledge search. It comprises 1,000 high‑resolution (4K+) images covering eight diverse daily‑life scenarios. To ensure exceptional quality, every image–question–answer triplet is meticulously crafted through collaboration between domain experts (PhDs from top‑tier global universities across diverse disciplines).

\begin{figure}
    \includegraphics[width=\textwidth]{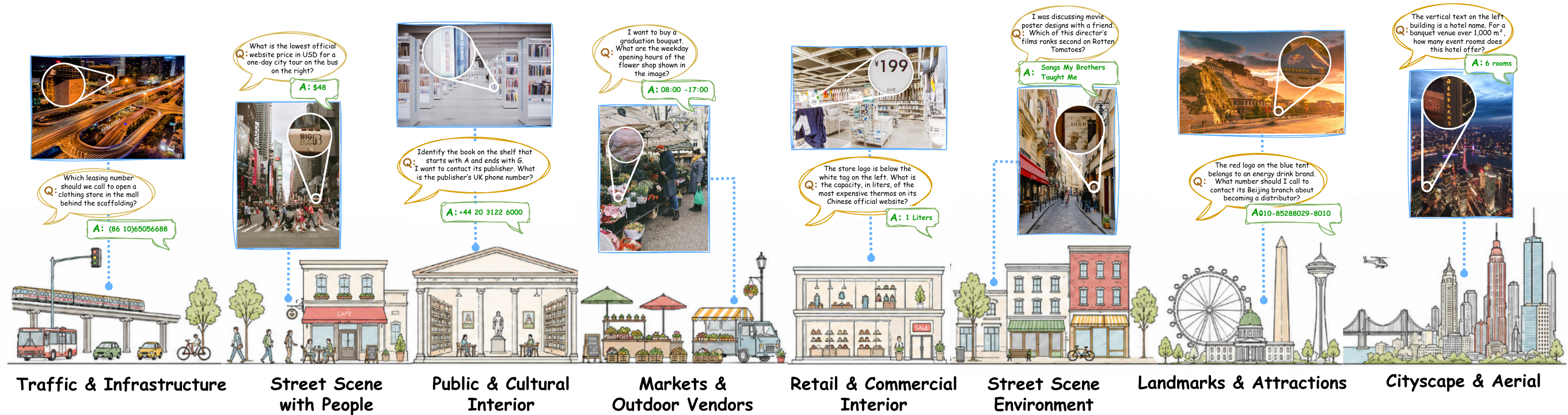}
    \captionof{figure}{\textbf{The Scenarios of Pix2Fact.} Detailed examples can be found in Appendix~\ref{appendix:figs}.}
    \vspace{-15pt}
    \label{fig:teaser}
\end{figure}

To this end, we construct Pix2Fact using a cohesive methodology that ensures high visual quality and explicitly demands web search for external knowledge. The process starts with a three‑stage image curation pipeline. First, high‑resolution images are sourced from license‑free platforms for legal compliance and diversity. Next, an automated pre‑screening step filters out any image below heuristic thresholds (e.g., resolution and file size), to maintain high‑fidelity imagery. Finally, PhD‑level experts manually discard those with poor composition or irrelevant content. To turn curated images into a benchmark, our annotation process follows two principles: (1) answers must integrate fine‑grained visual details, and (2) each question must require web search. Concretely, annotators first identify a visually grounded detail (such as a price in US dollars on a bar menu), and naturally extend it into a question that demands open‑web search, e.g., “What is the equivalent price in British pounds at the current exchange rate?” Such a question mirrors what a British tourist would genuinely ask. It arises organically from the visual context, requiring the model to both locate the fine detail (the dollar amount) and perform a web search (fetch the exchange rate and convert). Thus, each Pix2Fact question serves as a genuine test of precise visual grounding combined with effective knowledge retrieval.

The evaluation of ten leading VLMs (including Gemini-3.1-Pro and GPT-5.4) on Pix2Fact reveals a striking result: even with access to visual ground truth and search tools, the most advanced model achieves only 51.7\% average accuracy. Our analysis identifies the lack of external knowledge as the primary bottleneck. However, the fact that providing external knowledge via search tools still leaves accuracy capped at 51.7\% indicates that the challenge is not merely knowledge access. Instead, models also struggle with precise visual grounding and knowledge search, i.e., the very conjunction that Pix2Fact is designed to test. Our analysis attributes this low accuracy to three factors: frequent visual grounding errors even with visual ground truth, shallow search execution (models perform only a few rounds of keyword trials without iterative refinement or re‑searching), and VLM's inability to retrieve long‑tail, unstructured local information, due to poor query formulation, ineffective multi‑step search, and failure to extract relevant details from web pages that are not directly indexed or require manual navigation (e.g., small business hours, local signage, obscure phone numbers, temporary events, vendor details, or counting Bank of America branches offering home loans).

This significant gap underscores a critical limitation in current vision‑language models. They cannot reliably link fine‑grained visual details with relevant external knowledge through search. Therefore, Pix2Fact establishes an essential benchmark to drive progress toward multimodal systems that integrate fine‑grained perception with robust knowledge search. By explicitly targeting this conjunction, the dataset provides a meaningful testbed for developing and diagnosing more sophisticated, knowledge‑aware vision‑language architectures.

\begin{figure*}
    \centering
    \includegraphics[width=1\textwidth]{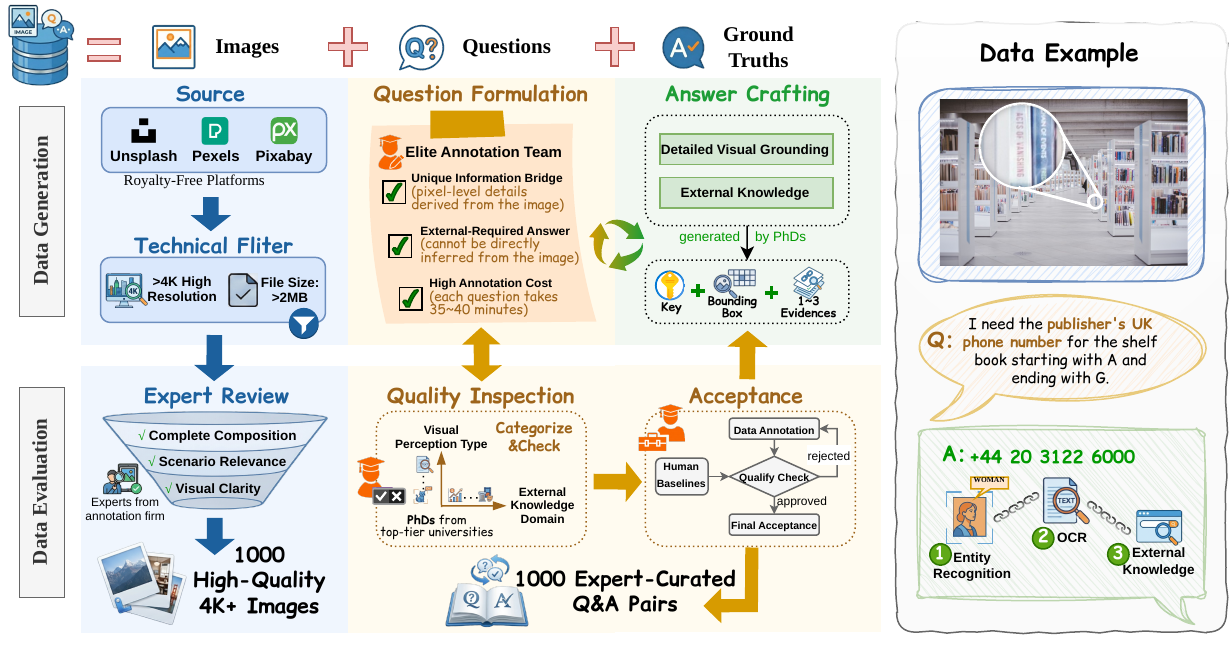}
    \captionof{figure}{\textbf{The Process of Constructing Pix2Fact Benchmark.}}
    \label{fig:constructing_pix2fact}
\end{figure*}

\section{Related Work}

\textbf{Detailed Visual Grounding Benchmark}
Recent visual grounding benchmarks have significantly advanced fine-grained perception in complex domains such as ultra-high-resolution aerial imagery, computational pathology, remote sensing, and 3D scene understanding, often by scaling to gigapixel scenes or incorporating novel sensors. A prevalent and limiting trend in these benchmarks, however, is their treatment of the perceptual localization task in isolation from knowledge-intensive, multi-step knowledge search. For instance, while GigaGrounding~\cite{Ma_2024_CVPR} introduces multi-hop referring expressions for gigapixel aerial imagery, its multi-step knowledge search is confined to spatial and visual cues within the scene itself. Benchmarks in remote sensing (e.g., AerialVG~\cite{Liu_2025_ICCV}, VRSBench~\cite{NEURIPS2024_05b7f821}) and 3D scenes (e.g., Anywhere3D-Bench~\cite{wang2025from}) rigorously evaluate the understanding of visible content and geometric relationships but do not incorporate external world knowledge. Even in specialized domains like medicine, where PathVG~\cite{ZhoChu_PathVG_MICCAI2025} and MedSG-Bench~\cite{yue2025medsg} integrate domain-specific terminology, the task remains bound to the immediate visual context without engaging broader knowledge retrieval. This limitation persists in benchmarks like LuoJia-VG~\cite{SHE2025104706}, which frame ``scene knowledge'' strictly as contextual facts intrinsic to the imagery rather than as connections to an open-world knowledge base. Similarly, search mechanisms such as $V^*$~\cite{wu2024v}, designed for detailed queries in everyday scenes, remain constrained by internal priors without access to open-world knowledge. We argue that integrating perception with open-world knowledge is a fundamental capability required for vision-language agents (VLAs) to transition from laboratory benchmarks to robust, real-world applications.

\textbf{VLM Reasoning Benchmark}
The release of models like DeepSeek-R1~\cite{guo2025deepseek} has intensified focus on evaluating complex reasoning in large language models~\cite{wu2025reasoning,lin-etal-2024-criticbench,dong2025textttclrbench,zhu2024dyval}, spurring the creation of multimodal benchmarks that establish rigorous frontiers for multi-hop reasoning across knowledge graphs~\cite{nguyen2024direct}, textual events~\cite{saha2025learning}, and visual data~\cite{zhao2024benchmarking}. However, these benchmarks predominantly address individual dimensions of the challenge, i.e., excelling either at external knowledge integration or fine-grained visual reasoning, but fail to assess the synergistic combination essential for advanced real-world applications. For instance, while benchmarks like ReasonVQA~\cite{tran2025reasonvqa} construct multi-hop questions over knowledge graphs, they decouple this symbolic inference from challenging visual understanding. Conversely, benchmarks such as MathSearch~\cite{madanmath} and SlideVQA~\cite{tanaka2023slidevqa} expertly couple visual interpretation with sequential reasoning but operate in a closed-world context without external knowledge retrieval. This divide persists elsewhere. MEQA~\cite{li2024meqa} advances multi-hop reasoning through dynamic event chains but remains purely textual, while multi-image benchmarks like MIRB~\cite{DBLP:journals/corr/abs-2406-12742} and MuirBench~\cite{wang2025muirbench} focus on perceptual reasoning within provided image sets rather than on active retrieval from expansive knowledge sources. Consequently, a critical gap exists for a benchmark that jointly entails fine-grained visual grounding and multi-hop active knowledge retrieval, which we believe is an integrated capability fundamental for the next generation of expert-level, knowledge-aware vision-language systems.

\textbf{Retrieval Augmented Generation Benchmark for VLM}
Knowledge-intensive tasks have been extensively studied since the emergence of large language models (LLMs)~\cite{gao2023retrieval,tang2024multihoprag,yang2024crag}. Recent benchmarks and methods have begun integrating retrieval with vision-language tasks. VLR-Bench~\cite{lim2025vlr} contributes a multilingual retrieval evaluation setup, BOK-VQA~\cite{kim2024bok} introduces a large-scale knowledge-grounded V$(Q,A)$dataset with structured knowledge injection, and EchoSight~\cite{yan-xie-2024-echosight} validates an effective image-to-knowledge retrieval framework. Their core strengths lie in explicitly linking visual questions to external knowledge. However, a key shared limitation is their reliance on simplified settings, such as closed-domain retrieval or coarse visual analysis, which fails to address the need for models that jointly perform fine-grained visual understanding and open-world knowledge retrieval.

To further distinguish the difference between Pix2Fact and other existing ones, we elaborate the benchmark details in Table~\ref{tab:benchmark_comparison}. While earlier benchmarks excel at evaluating isolated capabilities, such as fine-grained visual grounding, multi-hop reasoning, or knowledge-augmented retrieval, they rarely assess the synergistic combination of these skills required for expert-level multimodal understanding. Pix2Fact uniquely demands that a model: (1) perform fine-grained visual analysis to pinpoint a precise detail within a complex scene; and (2) use that detail to actively query open-world knowledge. This design moves beyond closed-world, perception-only, or coarsely prompted evaluation paradigms, and directly targets the integrated perception, retrieval, and knowledge search abilities essential for next-generation robust vision-language systems.

\begin{figure}[!t]
    \centering
    \begin{subfigure}[b]{0.3\textwidth}
        \centering
        \includegraphics[height=3.5cm]{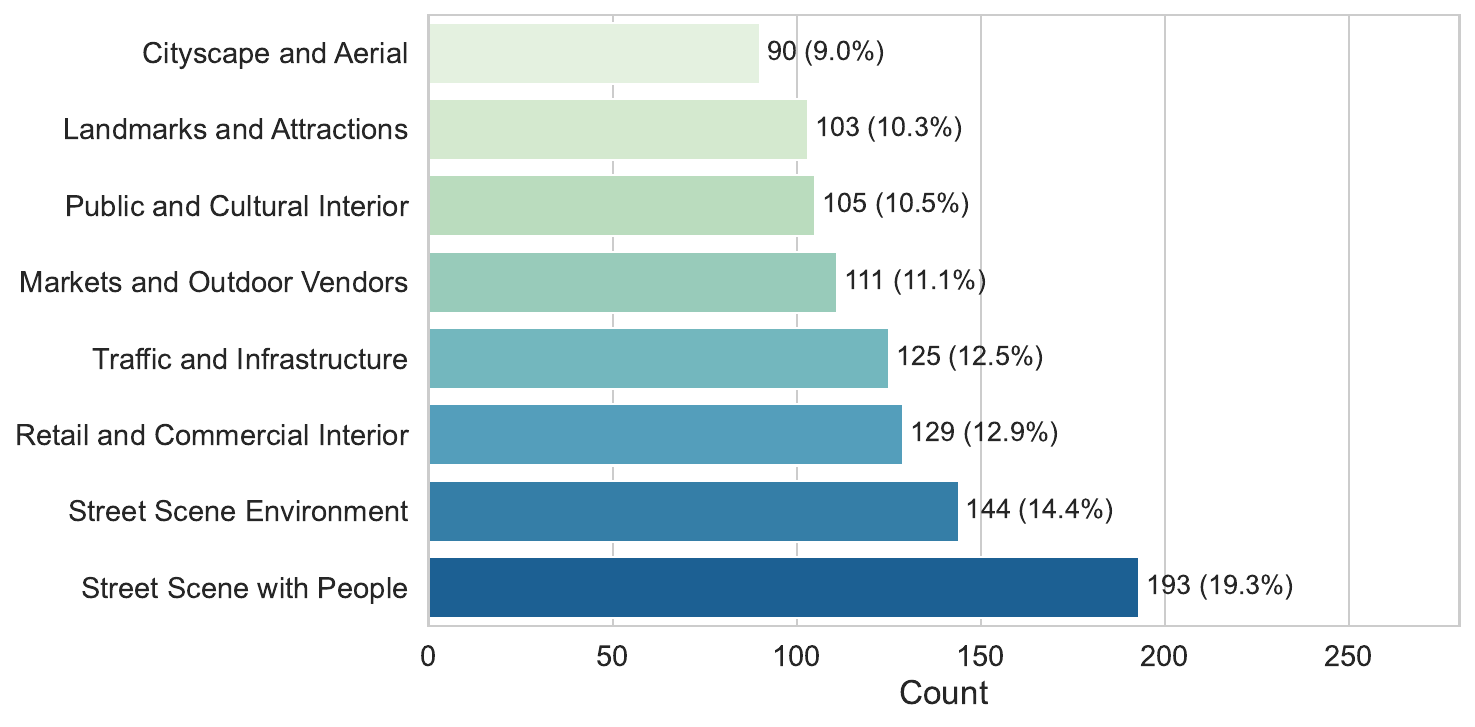}
        \caption{Image categories.}
    \end{subfigure}
    \hspace{100pt}
    \begin{subfigure}[b]{0.3\textwidth}
        \centering
        \includegraphics[height=3.5cm]{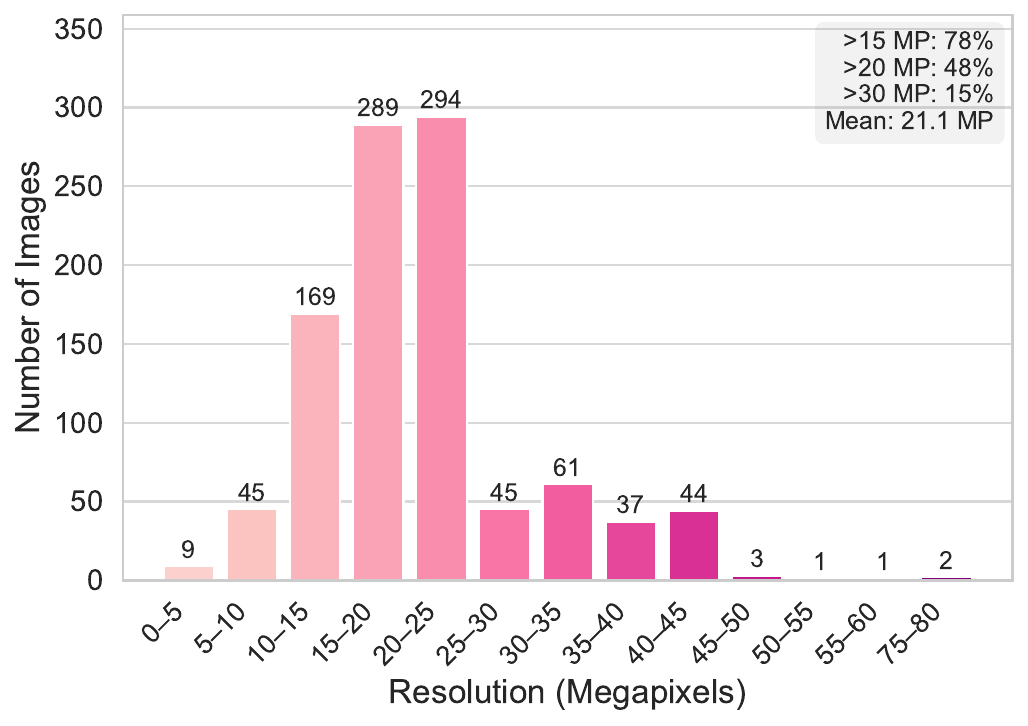}
        \caption{Resolution distribution.}
        \label{fig:resolution_distribution}
    \end{subfigure}
    \caption{\textbf{Statistics of the Pix2Fact Dataset.}
    (a) The 1,000 questions are distributed across 8 urban scene categories, ranging from 90 to 193 samples per category, indicating a balanced distribution without severe class bias.
    (b) The majority of images are high-resolution: 78\% exceed 15~MP and 48\% exceed 20~MP, with a mean resolution of 21.1~MP, ensuring rich visual detail for fine-grained grounding tasks.}
    \label{fig:stats_fix2fact}
\end{figure}

\section{Pix2Fact}

\subsection{An Overview of Pix2Fact}
We present Pix2Fact, a novel visual question-answering benchmark designed to evaluate the fine-grained visual grounding and deliberate, knowledge‑intensive search of vision‑language models in daily‑life scenarios. It comprises 1,000 high‑resolution (4K+) images across eight diverse real‑world categories, including streets with pedestrians, street environments, retail \& commercial interiors, traffic \& infrastructure, outdoor markets \& vendors, public \& cultural interiors, landmarks \& attractions, and cityscapes \& aerial views, ensuring broad applicability. Pix2Fact challenges models to integrate precise visual evidence with structured world knowledge, moving beyond simple recognition toward a deeper scene understanding. Dataset statistics and category coverage are detailed in Figure~\ref{fig:stats_fix2fact}.

To transform the images into a benchmark, Pix2Fact provides 1,000 expert-curated $(Q, A)$ pairs (one per high-resolution image). All questions and corresponding ground-truth answers were meticulously crafted by PhD researchers from top-tier global universities, ensuring exceptional quality and rigor. These domain experts provided the necessary expertise to design and validate complex, knowledge-intensive questions that require both fine-grained visual grounding and deliberate web search. The annotation workflow consisted of three defined phases: initial $(Q, A)$ crafting, multi-stage quality inspection, and final acceptance, ensuring that every question meets our strict criteria, requiring unique visual grounding and external knowledge. This rigorous, expert-driven process guarantees the benchmark's consistency and establishes a high-quality standard for evaluating fine-grained visual grounding together with knowledge search. For comprehensive access to our full dataset, please visit our data \href{https://huggingface.co/datasets/anon-submission-11235/Pix2fact}{here}. We also suggest to use our \href{https://anonymous.4open.science/r/pix2fact_eval-5C85/README.md}{benchmark test suite} for the convenience of reproduction.


\subsection{Image Data Curation Process}

This procedure outlines a clear, two-stage methodology for curating high-quality images. It begins with sourcing photographs exclusively from approved, royalty-free platforms (Unsplash\footnote{Unsplash: Please visit \textcolor{blue}{https://unsplash.com/} for details.}, Pexels\footnote{Pexels: Please visit \textcolor{blue}{http://www.pexels.com/} for details.}, and Pixabay\footnote{Pixabay: please visit \textcolor{blue}{https://pixabay.com/} for details.}) ensuring all materials are legally cleared for professional application. Each candidate image is then evaluated against stringent criteria. First, the required technical specifications stipulate a minimum file size of 2MB, and the resolution must be high (within the 4K to 8K range) while retaining sharpness and clarity even upon close inspection. Second, it must meet visual standards, meaning it should be well-composed, free of watermarks or distortion, and feature a strong, unobstructed subject. The selection process unfolds through two distinct, complementary phases designed to balance efficiency with qualitative rigor. The first phase is an automated technical screening, where images are rapidly filtered using specialized software to eliminate any files that fail to meet the fundamental size ($\geq2$ MB) and resolution (4K-8K) benchmarks. This initial gate ensures that only technically viable candidates proceed. Those that pass advance to the decisive second phase, i.e., a detailed manual review by the human expert. In this stage, an expert carefully examines each image. Any images that are deemed poorly composed, unclear, or thematically misaligned being systematically excluded. This combined approach results in a final collection of images that are not only technically excellent and copyright-safe, but also visually effective and ready for immediate annotation.

Our final dataset comprises ultra-high-resolution images, as detailed in Figure~\ref{fig:resolution_distribution}. In total, 78\% of images exceed 15 megapixels and 48\% exceed 20 megapixels, with a mean resolution of 21.1~MP and a maximum of 77.7~MP. Notably, 583 images (58\%) fall within the 15 to 25~MP range. The prevalence of such high-fidelity images provides the intricate visual detail necessary for generating accurate and complex queries in visual grounding tasks.

\definecolor{lightblue}{RGB}{230,240,255}
\definecolor{lightorange}{RGB}{255,240,230}
\definecolor{lightgrey}{RGB}{240,240,240}
\definecolor{lightgreen}{RGB}{230,255,230}
\definecolor{darkgreen}{RGB}{0,100,50}  

\newcommand{\phdicon}{\includegraphics[height=0.8em]{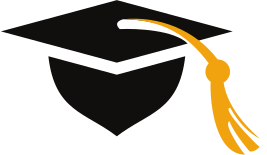}}
\newcommand{\geminiicon}{\includegraphics[height=0.8em]{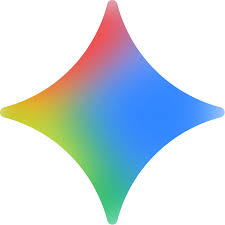}}
\newcommand{\searchicon}{\includegraphics[height=0.8em]{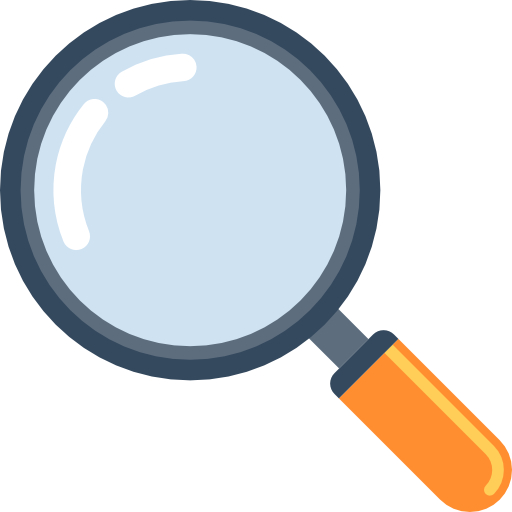}}

\subsection{$(Q,A)$ Pair Construction Process}

The annotation process follows a carefully structured three-tier quality control system. Each question is authored by a doctoral-level annotator, undergoes a full review by a second doctoral reviewer, and is finalized by a senior doctoral expert with professional quality inspection experience. If a question does not pass review at any stage, it is returned to the annotator for revision or reconstruction. This multi-layered review, conducted entirely by qualified researchers, ensures the foundational reliability and academic quality of each $(Q,A)$ pair.

The design of each question follows two fundamental principles. First, the vision clue, i.e., the specific search keyword, must be derived exclusively from fine-grained, pixel-level details visible in the image, requiring models to perform localized visual analysis. Second, the final answer must be grounded in external, verifiable knowledge that cannot be inferred from the image alone, thereby compelling the model to conduct a knowledge search. Crafting a single question that satisfies both conditions is a demanding and deliberate process, costing an average of 35 to 40 minutes per validated item. To effectively evaluate model performance during visual grounding and open-world knowledge retrieval, we design questions that explicitly require web search. This design ensures both question complexity and a meaningful assessment of search capacity. We define external knowledge search as logical chains involving relationships among three or more entities, or tasks that demand comparative analysis, enumeration, or commonsense inference across multiple objects or image regions. The details on the protocols and SOP to guarantee the high quality of $(Q,A)$ can be found in Appendix~\ref{qa_construction_detail}.

\vspace{-10pt}
\section{Benchmarking SOTA VLMs}

\definecolor{lightblue}{RGB}{230,240,255}
\definecolor{lightorange}{RGB}{255,240,230}
\definecolor{lightgrey}{RGB}{240,240,240}
\definecolor{lightgreen}{RGB}{230,255,230}
\definecolor{gray0}{RGB}{255,255,255}
\definecolor{gray1}{RGB}{245,245,245}
\definecolor{gray2}{RGB}{235,235,235}
\definecolor{gray3}{RGB}{225,225,225}
\definecolor{gray4}{RGB}{215,215,215}
\definecolor{gray5}{RGB}{205,205,205}
\definecolor{gray6}{RGB}{195,195,195}
\definecolor{gray7}{RGB}{185,185,185}
\definecolor{darkgreen}{RGB}{0,100,50}

\begin{table*}[!t]

\begin{subtable}{\textwidth}
\centering
\scalebox{0.8}{
\begin{tabular}{lccccc}
\toprule
Models
  & \makecell{C1\\Orig / No Search\\(\%)}
  & \makecell{C2\\Orig / Search\\(\%)}
  & \makecell{C3\\Crop / No Search\\(\%)}
  & \makecell{C4\\Crop / Search\\(\%)}
  & \makecell{Avg\\(\%)} \\
\midrule
\rowcolor{lightblue}
\multicolumn{6}{l}{\textit{Open-weights models}} \\
\midrule
\includegraphics[height=0.8em]{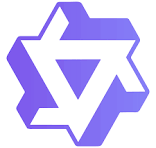} Qwen3.6-27B
  & \cellcolor{gray0}4.7
  & \cellcolor{gray2}16.8
  & \cellcolor{gray0}4.9
  & \cellcolor{gray4}26.2
  & \cellcolor{gray2}13.2 \\
\includegraphics[height=0.8em]{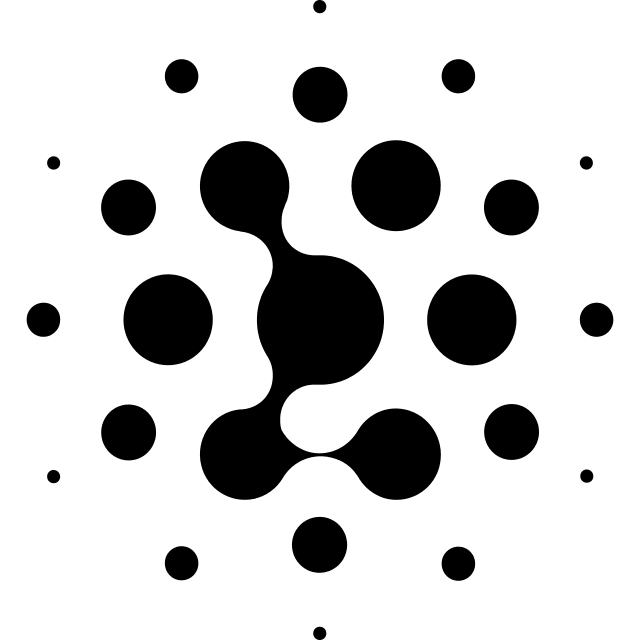} GLM-4.6V
  & \cellcolor{gray0}2.9
  & \cellcolor{gray1}11.4
  & \cellcolor{gray1}6.6
  & \cellcolor{gray3}23.9
  & \cellcolor{gray1}11.2 \\
\includegraphics[height=0.8em]{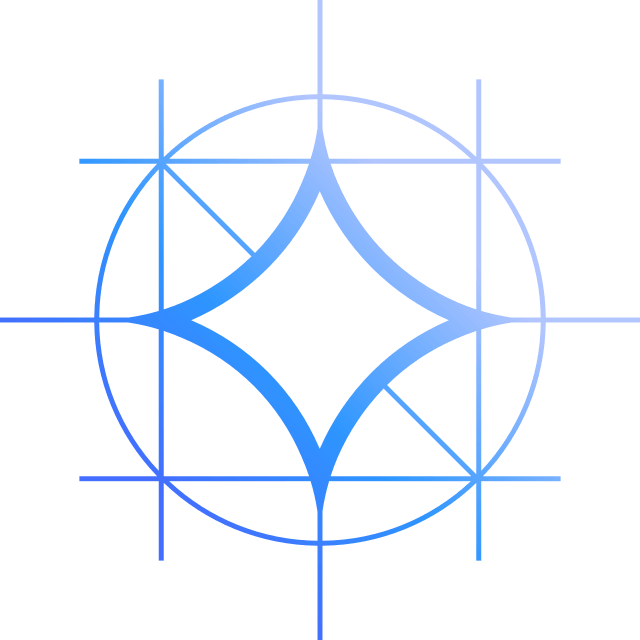} Gemma4-31B
  & \cellcolor{gray0}2.8
  & \cellcolor{gray1}8.2
  & \cellcolor{gray1}6.2
  & \cellcolor{gray3}19.6
  & \cellcolor{gray1}9.2 \\
\midrule
\rowcolor{lightorange}
\multicolumn{6}{l}{\textit{Closed-weights models}} \\
\midrule
\includegraphics[height=0.8em]{fig/gemini.jpg} gemini-3.1-pro
  & \cellcolor{gray3}\textcolor{darkgreen}{\textbf{18.4}}
  & \cellcolor{gray6}\textcolor{darkgreen}{\textbf{42.4}}
  & \cellcolor{gray3}\textcolor{darkgreen}{\textbf{21.0}}
  & \cellcolor{gray7}\textcolor{darkgreen}{\textbf{51.7}}
  & \cellcolor{gray5}\textcolor{darkgreen}{\textbf{33.4}} \\
\includegraphics[height=0.8em]{fig/gemini.jpg} gemini-2.5-pro
  & \cellcolor{gray2}14.6
  & \cellcolor{gray4}29.1
  & \cellcolor{gray3}18.6
  & \cellcolor{gray6}39.0
  & \cellcolor{gray4}25.3 \\
\includegraphics[height=0.8em]{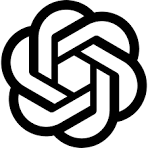} GPT-5.4
  & \cellcolor{gray1}8.5
  & \cellcolor{gray2}17.9
  & \cellcolor{gray2}14.5
  & \cellcolor{gray5}32.9
  & \cellcolor{gray3}18.5 \\
\includegraphics[height=0.8em]{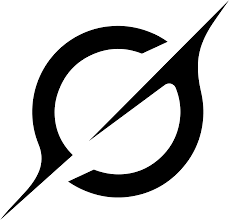} Grok-4.20
  & \cellcolor{gray0}4.4
  & \cellcolor{gray3}22.3
  & \cellcolor{gray1}7.3
  & \cellcolor{gray6}38.8
  & \cellcolor{gray2}18.2 \\
\includegraphics[height=0.8em]{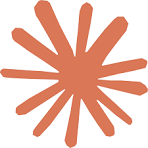} Claude-Opus-4.7$^{*}$
  & \cellcolor{gray2}13.3
  & \cellcolor{gray0}\textit{N/A}
  & \cellcolor{gray2}16.3
  & \cellcolor{gray0}\textit{N/A}
  & \cellcolor{gray2}14.8 \\
\includegraphics[height=0.8em]{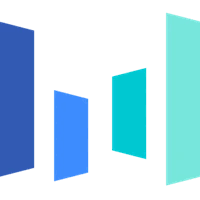} Doubao-2.0
  & \cellcolor{gray1}8.0
  & \cellcolor{gray1}8.6
  & \cellcolor{gray2}12.5
  & \cellcolor{gray2}15.1
  & \cellcolor{gray1}11.1 \\
\includegraphics[height=0.8em]{fig/doubao.png} Doubao-1.8
  & \cellcolor{gray1}7.2
  & \cellcolor{gray1}10.6
  & \cellcolor{gray1}8.4
  & \cellcolor{gray2}17.2
  & \cellcolor{gray1}10.9 \\
\bottomrule
\end{tabular}
}
\end{subtable}

\vspace{6pt}

\begin{subtable}{\textwidth}
\centering
\scalebox{0.8}{
\begin{tabular}{lccccccr}
\toprule
Models
  & \makecell{$\Delta V_{\text{ns}}$\\\textit{\small C3$-$C1}}
  & \makecell{$\Delta V_{\text{ws}}$\\\textit{\small C4$-$C2}}
  & \makecell{$\Delta S_{\text{orig}}$\\\textit{\small C2$-$C1}}
  & \makecell{$\Delta S_{\text{crop}}$\\\textit{\small C4$-$C3}}
  & \makecell{$\Delta$Total\\\textit{\small C4$-$C1}}
  & \makecell{C2$-$C3\\\textit{\small search$-$crop}}
  & \makecell{Synergy\\\textit{\small $\Delta V_{\text{ws}}/\Delta V_{\text{ns}}$}} \\
\midrule
\rowcolor{lightblue}
\multicolumn{8}{l}{\textit{Open-weights models}} \\
\midrule
\includegraphics[height=0.8em]{fig/qwen.png} Qwen3.6-27B
  & \cellcolor{gray0}$+$0.2
  & \cellcolor{gray2}$+$9.4
  & \cellcolor{gray3}$+$12.1
  & \cellcolor{gray5}$+$21.3
  & \cellcolor{gray5}$+$21.5
  & \cellcolor{gray3}$+$11.9
  & \cellcolor{gray0}$\infty^{\dagger}$ \\
\includegraphics[height=0.8em]{fig/zhipu.png} GLM-4.6V
  & \cellcolor{gray1}$+$3.7
  & \cellcolor{gray3}$+$12.5
  & \cellcolor{gray2}$+$8.5
  & \cellcolor{gray4}$+$17.3
  & \cellcolor{gray5}$+$21.0
  & \cellcolor{gray1}$+$4.8
  & \cellcolor{gray3}$3.38{\times}$ \\
\includegraphics[height=0.8em]{fig/gemma-color.png} Gemma4-31B
  & \cellcolor{gray1}$+$3.4
  & \cellcolor{gray3}$+$11.4
  & \cellcolor{gray1}$+$5.4
  & \cellcolor{gray3}$+$13.4
  & \cellcolor{gray4}$+$16.8
  & \cellcolor{gray1}$+$2.0
  & \cellcolor{gray3}$3.35{\times}$ \\
\midrule
\rowcolor{lightorange}
\multicolumn{8}{l}{\textit{Closed-weights models}} \\
\midrule
\includegraphics[height=0.8em]{fig/gemini.jpg} gemini-3.1-pro
  & \cellcolor{gray1}$+$2.6
  & \cellcolor{gray2}$+$9.3
  & \cellcolor{gray5}\textcolor{darkgreen}{\textbf{$+$24.0}}
  & \cellcolor{gray7}$+$30.7
  & \cellcolor{gray7}$+$33.3
  & \cellcolor{gray5}\textcolor{darkgreen}{\textbf{$+$21.4}}
  & \cellcolor{gray4}$3.58{\times}$ \\
\includegraphics[height=0.8em]{fig/gemini.jpg} gemini-2.5-pro
  & \cellcolor{gray1}$+$4.0
  & \cellcolor{gray2}$+$9.9
  & \cellcolor{gray3}$+$14.5
  & \cellcolor{gray5}$+$20.4
  & \cellcolor{gray5}$+$24.4
  & \cellcolor{gray3}$+$10.5
  & \cellcolor{gray1}$2.47{\times}$ \\
\includegraphics[height=0.8em]{fig/gpt.png} GPT-5.4
  & \cellcolor{gray2}\textcolor{darkgreen}{\textbf{$+$6.0}}
  & \cellcolor{gray3}$+$15.0
  & \cellcolor{gray2}$+$9.4
  & \cellcolor{gray4}$+$18.4
  & \cellcolor{gray5}$+$24.4
  & \cellcolor{gray1}$+$3.4
  & \cellcolor{gray1}$2.50{\times}$ \\
\includegraphics[height=0.8em]{fig/grok.png} Grok-4.20
  & \cellcolor{gray1}$+$2.9
  & \cellcolor{gray4}\textcolor{darkgreen}{\textbf{$+$16.5}}
  & \cellcolor{gray4}$+$17.9
  & \cellcolor{gray7}\textcolor{darkgreen}{\textbf{$+$31.5}}
  & \cellcolor{gray7}\textcolor{darkgreen}{\textbf{$+$34.4}}
  & \cellcolor{gray3}$+$15.0
  & \cellcolor{gray7}\textcolor{darkgreen}{\textbf{$5.69{\times}$}} \\
\includegraphics[height=0.8em]{fig/claude.png} Claude-Opus-4.7$^{*}$
  & \cellcolor{gray1}$+$3.0
  & \cellcolor{gray0}\textit{N/A}
  & \cellcolor{gray0}\textit{N/A}
  & \cellcolor{gray0}\textit{N/A}
  & \cellcolor{gray0}\textit{N/A}
  & \cellcolor{gray0}\textit{N/A}
  & \cellcolor{gray0}\textit{N/A} \\
\includegraphics[height=0.8em]{fig/doubao.png} Doubao-2.0
  & \cellcolor{gray1}$+$4.5
  & \cellcolor{gray2}$+$6.5
  & \cellcolor{gray0}$+$0.6
  & \cellcolor{gray1}$+$2.6
  & \cellcolor{gray2}$+$7.1
  & \cellcolor{gray0}$-$3.9
  & \cellcolor{gray0}$1.44{\times}$ \\
\includegraphics[height=0.8em]{fig/doubao.png} Doubao-1.8
  & \cellcolor{gray1}$+$1.2
  & \cellcolor{gray2}$+$6.6
  & \cellcolor{gray1}$+$3.4
  & \cellcolor{gray2}$+$8.8
  & \cellcolor{gray2}$+$10.0
  & \cellcolor{gray1}$+$2.2
  & \cellcolor{gray7}$5.50{\times}$ \\
\midrule
\textbf{9-model Avg.}
  & \cellcolor{gray1}$\mathbf{+3.2}$
  & \cellcolor{gray3}$\mathbf{+10.8}$
  & \cellcolor{gray3}$\mathbf{+10.6}$
  & \cellcolor{gray4}$\mathbf{+18.3}$
  & \cellcolor{gray5}$\mathbf{+21.4}$
  & \cellcolor{gray2}$\mathbf{+7.5}$
  & \cellcolor{gray3}$\mathbf{3.49{\times}}$ \\
\bottomrule
\end{tabular}
}
\end{subtable}

\caption{\textbf{VLM Evaluation under Four Conditions and Gain Decomposition.}
\textbf{(a) Performance under C1--C4.}
C1: original image, no web search; C2: original image, with web search; C3: expert-cropped visual ground truth, no web search; C4: expert-cropped visual ground truth, with web search.
Avg is the mean accuracy (\%) over C1--C4.
\textbf{(b) Gain decomposition.}
$\Delta V_{\text{ns}}$ and $\Delta V_{\text{ws}}$ measure the gain from expert cropping without and with search, respectively.
$\Delta S_{\text{orig}}$ and $\Delta S_{\text{crop}}$ measure the gain from enabling search on original and cropped images.
$\Delta$Total is the overall gain from C1 to C4.
C2$-$C3 contrasts search advantage against cropping advantage (positive = search helps more).
Synergy ($\Delta V_{\text{ws}}/\Delta V_{\text{ns}}$) measures how much search amplifies the cropping gain.
A heatmap uses a gradient from white (lowest) to dark gray (highest) for quick comparison.
\textcolor{darkgreen}{\textbf{Bold green}} indicates the highest score in each column.
$^{\dagger}$Qwen3.6-27B has $\Delta V_{\text{ns}}\approx 0$, making the ratio undefined.
$^{*}$Claude-Opus-4.7 does not support search-enabled conditions (C2, C4).}
\vspace{-15pt}
\label{tab:combined}
\end{table*}

\subsection{Experiment Setups}
\textbf{Proprietary Models Configurations}. 
During inference, a temperature of 0 was employed for all Vision-Language Models (VLMs) to ensure deterministic outputs. An exception was made for GPT-5.4, as its official API constrains the temperature parameter to a value of 1.
For reasoning models such as Gemini-3.1-Pro, GPT-5.4, Claude Opus 4.7, and Doubao-2.0-pro, 
we configure the inference to utilize the maximum permissible reasoning token limit.
To ensure reproducibility and a fair comparison, the web-search functionality is implemented using the official code released by the respective model and service providers.
For instance, we enable the native Google search tool for Gemini-2.5/3.1-Pro allowing the model to invoke this tool and retrieve search responses during its reasoning process. 
For GPT-5.4, we use the responses API\footnote{Please visit \textcolor{blue}{https://platform.openai.com/docs/api-reference/responses} for details.} with the native web search tool. 
All web search tools are configured with default settings, enabling the model to decide whether to perform zero or multiple web search calls within the context limit.  
For models like Claude, where search functionality requires client-side implementation, the web-search capability is omitted from our evaluation.  
Some model APIs impose constraints on maximum image file size and pixel count. For instance, Doubao limits images to under 10MB and 36 million pixels. When these limits are exceeded, we dynamically resize the input image, preserving its aspect ratio, to meet the specified constraints.
The detailed VLMs API configurations can be found in Appendix~\ref{appendix:api_config}. After three failed API call attempts, an answer is auto-flagged as False. This affects $<5\%$ of data per model.

\textbf{Open-weights Models Configurations}. 
We evaluated Qwen3.6-27B, Gemma4-31B-it, and GLM-4.6V on the Pix2fact benchmark. During inference, we set the temperature to 0 to ensure reproducibility. To ensure a fair comparison, we standardized the context window limit at 128K tokens for all open-weights models. Additionally, image inputs were capped at 15 MP through aspect-ratio-preserving resizing.

\textbf{Search Agent Implementation}. 
To enable open-weights models to search and read external web sources, we implemented an agent framework following the ReAct \cite{yao2023iclr-react} paradigm. The agent is prompted to generate a reasoning process regarding tool selection, followed by emitting one or more tool calls with specific parameters. Next, our agent framework executes the selected tools and retrieves the responses. The prompt for the search agent can be found in Appendix \ref{appendix:search_agent_prompt}. The agent utilizes these tools iteratively to resolve the given instruction. We equipped all open-source models with the following tools: \textbf{(1). A web-search tool}, powered by the You.com Search API\footnote{Please visit \textcolor{blue}{https://api.you.com} for details.}, which returns a list of webpage summaries and their corresponding URLs based on a given query. \textbf{(2). A webpage-visit tool}, utilizing the Jina Reader API\footnote{Please visit \textcolor{blue}{https://jina.ai/reader} for details.} to convert raw webpage HTML into an LLM-friendly Markdown format based on the URLs retrieved from the search tool. \textbf{(3). A terminate tool}, a special mechanism allowing the model to either submit a final answer or declare its inability to address the query. This tool also outputs search plan and search query for later analysis. To handle the constraint of a 128,000-token context window, we prompt the agent to synthesize an answer based on the acquired information whenever the context limit is reached.

\textbf{LLM as Judge For Answer Validation}. 
We adopt an LLM-as-judge protocol to evaluate answer correctness, using Gemini-3.1-Pro and GPT-5.4 as independent judges. The detailed evaluation prompt is provided in Appendix~\ref{appendix:judge_prompt}. We observe that LLM judges are not always reliable: they tend to make errors when handling answer variants (e.g., phone numbers or addresses with different formats), complex answers, and answers that extend beyond the ground truth yet remain valid under human judgment. To assess judge reliability, we conduct two rounds of human-verified auditing on sampled model outputs. In the first round, using Gemini-3.1 outputs as evaluation targets, Gemini-3.1 achieves higher agreement with human annotations (94.1\%) compared to GPT-5.4 (91.1\%), with GPT-5.4 exhibiting a tendency toward stricter judgments (i.e., higher false negative rates). To further rule out potential bias toward outputs from models of the same provider, the second round evaluates Claude 4.7-generated answers, where Gemini-3.1 still maintains superior alignment (97.0\% vs. 94.7\%).

Based on the results, we select Gemini-3.1-Pro as the primary judge. We additionally compute inter-judge agreement over the full evaluation set (40,000 instances), obtaining a raw agreement of 95.86\% and a Cohen's $\kappa$ of 0.836, which indicates near-perfect agreement under low base-rate conditions. These findings demonstrate that with our evaluation prompt, the judge model can provide stable and reliable evaluation signals for our benchmark.

\subsection{Main Results}


To systematically evaluate the contributions of visual grounding and external knowledge retrieval, we design four evaluation conditions (C1--C4) by crossing two binary axes: image input type and web-search availability. For the image input axis, \textit{orig} denotes the original scene image, while \textit{crop} denotes an expert-annotated tight crop that isolates the target object or region, substantially reducing visual ambiguity. For the search axis, models either operate without external retrieval (\textit{no search}) or are permitted to issue web queries during inference (\textit{with search}). \textbf{C1 (orig / no search)} serves as the baseline, reflecting purely vision-driven recognition under realistic conditions. \textbf{C2 (orig / with search)} introduces retrieval while preserving the full image, testing whether search can compensate for visual difficulty. \textbf{C3 (crop / no search)} removes visual ambiguity via expert cropping while disabling external knowledge, isolating the model's intrinsic recognition capacity. \textbf{C4 (crop / with search)} combines both advantages and represents the best-case condition (upper bound).

\textbf{Key findings from C1 to C4.} 
Table~\ref{tab:combined} shows a consistent hierarchy: all models improve from C1 (original, no search) to C4 (cropped, with search), but with large differences in sensitivity. Under C1, Gemini-3.1-Pro already leads open-weight models by a factor of $4\times$. Web search (C2) provides the largest gain for top models (e.g., Gemini-3.1-Pro jumps from 18.4\% to 42.4\%), while cropping alone (C3) yields only modest improvement ($+3.2\%$ on average). The full combination C4 sets the ceiling at 51.7\% (Gemini-3.1-Pro), with Grok-4.20 showing the most dramatic recovery ($+34.4$ pp). Open-weight models lag persistently, with C4 averages of 19--26\% vs. 33--52\% for closed-weight leaders. Notably, even under optimal conditions, no model exceeds 51.7\%, confirming that Pix2Fact remains genuinely difficult despite search and visual ground truth.

\textbf{Key findings from gain decomposition.} 
The analysis in Table~\ref{tab:combined} reveals that web search is the dominant performance driver across nearly all models. On average, search contributes $\Delta S_{\text{orig}} = +10.6\%$ and $\Delta S_{\text{crop}} = +18.3\%$, whereas expert cropping alone yields only $\Delta V_{\text{ns}} = +3.2\%$. Crucially, search and cropping are strongly synergistic: the average synergy ratio of $3.49\times$ across nine models means that expert cropping triples in value when combined with web search, a pattern consistent across every model with available data. At the frontier, Grok-4.2 achieves the highest total gain ($\Delta\text{Total} = +34.4\%$) and the strongest search--crop synergy ($5.69\times$). Gemini-3.1-Pro leads in raw search utilization ($\Delta S_{\text{orig}} = +24.0\%$; C2$-$C3$ = +21.4\%$), confirming that its retrieval capability is uniquely powerful on full-resolution images. GPT-5.4 is the outlier in the cropping dimension: it gains the most from expert cropping alone ($\Delta V_{\text{ns}} = +6.0\%$), yet its comparatively lower synergy ($2.50\times$) suggests diminishing returns when search is added on top. By contrast, Doubao-2.0 is largely unresponsive to search: $\Delta S_{\text{orig}} = +0.6\%$ and a negative C2$-$C3 value of $-3.9\%$ indicate that cropping outweighs search for this model, exposing a fundamental limitation in its retrieval integration. Qwen3.6-27B presents a complementary failure mode: expert cropping without search yields virtually no gain ($\Delta V_{\text{ns}} = +0.2\%$), yet when combined with search it recovers to $\Delta V_{\text{ws}} = +9.4\%$, implying that its visual understanding is only unlocked when retrieval provides external context. Open-weight models achieve moderate total improvements ($\Delta\text{Total} \approx 17\%$--$22\%$) but remain below the closed-weight frontier, with lower synergy ratios ($3.35$--$3.38\times$), suggesting architectural or training constraints in jointly exploiting cropping and search. Taken together, these findings indicate that achieving high performance on Pix2Fact requires both high-quality visual grounding \emph{and} effective web retrieval---neither factor alone is sufficient.

\subsection{Deep Analysis}We conduct a detailed error analysis on 121 cleaned evaluation cases (from an initial pool of 152, after removing 31 contaminated samples due to judge disagreement or annotation issues). This analysis reveals several non-trivial insights into the failure modes of state-of-the-art VLMs under our strongest setting (crop + search). 

\textbf{Model-specific bottlenecks under strong conditions.}
Even under the strongest setting (C4), models exhibit highly heterogeneous failure patterns. Gemini-3.1-Pro approaches saturation (92.5\%), with residual errors primarily caused by reasoning failures, where relevant evidence is successfully retrieved but the model fails to derive the correct conclusion. In contrast, Claude shows a dominant knowledge bottleneck (87\% of its errors), corresponding to cases where the model performs search correctly but fails to obtain the necessary information from external sources. GPT-5.4, meanwhile, suffers from a combination of knowledge and visual grounding issues. Qwen exhibits a distinct failure mode: in 63\% of its errors, it fails to invoke the search tool at all, indicating a breakdown in tool-use rather than query formulation.

\textbf{Effect of cropping: beyond resolution enhancement.}
After correcting for judge noise, we identify 18 genuine visual grounding failures. Cropping successfully resolves all of them. Notably, its benefit is not limited to improving resolution: 72.2\% of cases involve either low resolution or incorrect focus, suggesting that cropping also acts as an attention re-localization mechanism rather than merely enlarging visual details. In the following example: under the uncropped image condition, the model attends to a different but visually similar object and therefore fails to recognize the sneakers’ logo as New Balance. Under the cropped condition, however, the model attends to the correct object and successfully identifies the Nike swoosh logo on the sneakers.

\begin{figure}[h]
\centering
\begin{tcolorbox}[colback=green!3!white,colframe=black,boxrule=0.8pt,arc=4pt,width=\linewidth]
{\sffamily
\small
\setstretch{1.12}
\raggedright
\textls[20]{%

Example Case Study \textit{index: 923-1}

\textbf{Ground Truth:} Nike

\textbf{Original Observation (\textcolor{red}{Incorrect}):} On the right side of the image, near a stone bollard, there is a short-haired woman with glasses wearing a purple patterned jacket and black pants. She
is wearing black sneakers with a distinctive \colorbox{rosePink!15}{white `N' logo} on the side, identifying the brand as \colorbox{rosePink!15}{New Balance}.

\textbf{Crop Observation (\textcolor{red}{Correct}):} In the image, the short-haired woman with glasses standing on the right is wearing black sneakers featuring a distinctive \colorbox{green!15}{white `swoosh'
logo}.

}
}
\end{tcolorbox}
\caption{\textbf{Case Study 923-1}: Crop-based correction of visual focus for accurate entity recognition.}
\label{prompt:user_generator_prompt_with_history}
\end{figure}
\vspace{-10pt}




\textbf{Search failures are dominated by upstream issues.}
Contrary to a common assumption, we find that knowledge integration is not the primary bottleneck. Only 17.4\% of failures arise from retrieved-but-unused knowledge. Instead, the majority of failures originate from earlier stages: query construction (43.5\%) and retrieval failure (39.1\%). This indicates that search-stage issues, rather than reasoning over retrieved content, dominate the error distribution.

\textbf{Systematic bias in model behavior.}
We observe a consistent behavioral pattern in Claude: instead of refusing to answer, it tends to produce confident but slightly incorrect responses (76.9\%), often involving near-miss factual errors. This contrasts with prior assumptions that closed models fail primarily due to over-conservative refusal. In the following example: Claude successfully identifies the correct brand and retrieves highly relevant evidence, but produces an approximately correct yet factually inaccurate final answer.  This example illustrates Claude's characteristic "confident-but-imprecise" failure pattern on long-tail factual knowledge.

\begin{figure}[h]
\centering
\begin{tcolorbox}[colback=green!3!white,colframe=black,boxrule=0.8pt,arc=4pt,width=\linewidth]
{\sffamily
\small
\setstretch{1.12}
\raggedright
\textls[20]{%

Example Case Study \textit{index: 415-1}

\textbf{Question:} For the well-known clothing brand appearing in the picture, how many stores did it have worldwide in the 2024 fiscal year?

\textbf{Ground Truth:} \colorbox{green!15}{2,495}

\textbf{Claude Comprehensive Answer:} \ldots According to Fast Retailing's (Uniqlo's parent company) annual report for fiscal year 2024 (ending August 31, 2024), Uniqlo had approximately
\colorbox{rosePink!15}{2,501} stores worldwide, including around 800 in Japan and over 1,700 internationally.

\textbf{Claude Final Answer (\textcolor{red}{Incorrect}):} Approximately \colorbox{rosePink!15}{2,501} stores

}
}
\end{tcolorbox}
\caption{\textbf{Case Study 415-1}: Confident but inaccurate numerical answer by Claude Opus 4.7.}
\label{fig:case_study_415}
\end{figure}
\vspace{-10pt}






\textbf{Residual failures beyond perception.}
Even after resolving visual grounding and enabling search, a large portion of errors persist. These are mainly due to (i) misuse of retrieved information (38.5\%) and (ii) retrieval failure (38.5\%), indicating that downstream reasoning and tool interaction remain critical challenges. This supports the claim that improvements in search pipelines yield larger gains than further visual enhancements. More detailed cases can be found in Appendix~\ref{appendix:failure_analysis}.

\vspace{-10pt}
\section{Conclusion}
\vspace{-6pt}
We introduce Pix2Fact, a benchmark for evaluating VLMs on the conjunction of fine-grained visual grounding and open-world knowledge search. Pix2Fact comprises 1,000 high-resolution images across eight daily-life scenarios, each paired with expert-crafted questions demanding both precise visual localization and deliberate web retrieval. Evaluating ten state-of-the-art VLMs reveals that even the best model (Gemini-3.1-Pro) achieves only 51.7\% accuracy under optimal conditions (cropped images with search), leaving a substantial performance gap. Our analysis attributes this shortfall to three factors, i.e., frequent visual grounding errors even with visual ground truth, shallow search execution, and inability to retrieve long-tail unstructured local information. Pix2Fact provides a critical testbed for driving progress toward multimodal systems that robustly integrate fine-grained perception with effective knowledge search.


\clearpage
\bibliographystyle{plain}
\bibliography{reference}


\appendix

\newpage

\section{Summary on Existing VQA Benchmarks}
\definecolor{lightgray}{RGB}{245,245,245}
\definecolor{lightblue}{RGB}{220,235,255}
\definecolor{lightorange}{RGB}{255,235,220}
\definecolor{lightgreen}{RGB}{230,255,230}
\newcommand{\cmark}{\textcolor{green!60!black}{\ding{51}}}  
\newcommand{\xmark}{\textcolor{red}{\ding{55}}}  
\newcommand{\wmark}{\textcolor{orange}{\ding{109}}}  
\begin{table*}[!h]
\centering
\small
\setlength{\tabcolsep}{8pt}
\scalebox{0.8}{
\begin{tabular}{l@{\hspace{16pt}}c@{\hspace{16pt}}c@{\hspace{16pt}}c}
\toprule
\textbf{Benchmark} & \makecell{\textbf{Detailed}\\\textbf{Grounding}} & \makecell{\textbf{RAG}\\\textbf{Use}} & \makecell{\textbf{Knowledge}\\\textbf{Intensity}} \\
\midrule
\rowcolor{lightblue}
\multicolumn{4}{l}{\textit{Detailed Visual Grounding Benchmarks}} \\
\midrule
GigaGrounding~\cite{Ma_2024_CVPR} & \cmark & \xmark & \xmark \\
PathVG~\cite{ZhoChu_PathVG_MICCAI2025} & \cmark & \xmark & \cmark \\
AerialVG~\cite{Liu_2025_ICCV} & \cmark & \xmark & \xmark \\
Anywhere3D-Bench~\cite{wang2025from} & \cmark & \xmark & \xmark \\
VRSBench~\cite{NEURIPS2024_05b7f821} & \cmark & \xmark & \xmark \\
MedSG-Bench~\cite{yue2025medsg} & \cmark & \xmark & \pmark \\
LuoJia-VG~\cite{SHE2025104706} & \cmark & \xmark & \pmark \\
$V^*$~\cite{wu2024v} & \cmark & \xmark & \xmark \\
\midrule
\rowcolor{lightorange}
\multicolumn{4}{l}{\textit{VLM Reasoning Benchmark}} \\
\midrule
ReasonVQA~\cite{tran2025reasonvqa} & \pmark & \pmark & \cmark \\
MEQA~\cite{li2024meqa} & \xmark & \xmark & \cmark \\
MathSearch~\cite{madanmath} & \cmark & \xmark & \xmark \\
MIRB~\cite{DBLP:journals/corr/abs-2406-12742} & \pmark & \xmark & \xmark \\
MuirBench~\cite{wang2025muirbench} & \pmark & \xmark & \xmark \\
SlideVQA~\cite{tanaka2023slidevqa} & \pmark & \xmark & \xmark \\
\midrule
\rowcolor{lightgreen}
\multicolumn{4}{l}{\textit{Retrieval Augmented Generation Benchmark for VLM}} \\
\midrule
VLR-Bench~\cite{lim2025vlr} & \pmark & \cmark & \cmark \\
BOK-VQA~\cite{kim2024bok} & \xmark & \cmark & \cmark \\
EchoSight~\cite{yan-xie-2024-echosight} & \xmark & \cmark & \cmark \\
\midrule
\rowcolor{lightgray}
Pix2Fact (This Work) & \cmark & \cmark & \cmark \\
\bottomrule
\end{tabular}
}
\caption{\textbf{Comparison of benchmarks.} \cmark~indicates full support, \xmark~indicates no support, and \pmark~indicates partial support.}
\vspace{-10pt}
\label{tab:benchmark_comparison}
\end{table*}

\section{LLM Prompt}\label{appendix:prompt}

\subsection{Instructions \& CoTs for answer generation:}
\begin{lstlisting}[
    style=codestyle,
    language={},
    basicstyle=\scriptsize\ttfamily,
    deletekeywords={input},
    keepspaces=true,
    columns=flexible,
    mathescape=false,
    literate={_}{\_}1,
    breaklines=true,
    breakindent=0pt,
    breakautoindent=false,
    postbreak=\space
]
You are a highly specialized AI designed to function as an automated visual analysis API. Your sole function is to analyze an image and a question provided by the user, and return your entire response as a single, valid JSON object.\n
--- RULES ---\n
Your entire output MUST be a single, valid JSON object.\n
Your response MUST start with { and end with }.\n
DO NOT output ANY text, explanations, apologies, or markdown formatting (like ```json) before or after the JSON object. Your response must be the raw JSON and nothing else.\n
The JSON object MUST contain these exact five keys: "Observation", "Search\ Plan", "Search Query", "Full Answer", and "Simple Answer". Adhere strictly to this schema.\n
--- KEY DEFINITIONS & SCHEMA ---\n
"Observation": (String) Describe specific visual details from the image URL relevant to the question.\n
"Search Plan": (List of Strings) Outline a step-by-step plan to find the necessary information online.\n
"Search Query": (List of Strings) Extract the exact search queries from your Search Plan.\n
"Comprehensive Answer": (String) Provide a comprehensive, final answer integrating observations and search results.\n
"Final Answer": (String) Provide only the core, direct answer. If a definitive, factual answer (e.g., a specific name, date, number) cannot be determined, you MUST output the exact string '[NO_DEFINITIVE_ANSWER]' in this field.\n
--- ONE-SHOT EXAMPLE ---\n
This is an example of a user request and your expected output.\n
User Input Example:\n
Image URL: <https://example.com/path/to/library_image.jpg>\n
Question: Who was the president of the USA when the book with 'Kjell' on its cover in the picture published?\n
Your Expected JSON Output Example:\n
{\n
"Observation": "On the top shelf of the book cart in the foreground, facing left, a book with a dark cover is visible. The author's name, "Kjell Westo," is printed in white, and below it is the title, "Hagring 38."",\n
"Search Plan": [\n
"Find the original publication date of the book titled "Hagring 38" by Kjell Westo.",\n
"Identify who was the President of the United States during the publication year of the book."\n
],\n
"Search Query": [\n
"Hagring 38 Kjell Westo publication date",\n
"who was US president in 2013"\n
],\n
"Comprehensive Answer": "The book visible in the image is "Hagring 38" by Kjell Westo, which was originally published in 2013. In that year, the president of the USA was Barack Obama, who was in his second term.",\n
"Final Answer": "Barack Obama"\n
}\n
--- YOUR TASK ---\n
Input Image: [Input your Image or Image URL]\n
Input Question: [Input your Quesiton]\n
\end{lstlisting}

\subsection{Instructions \& CoTs for search agent:}\label{appendix:search_agent_prompt}
\begin{lstlisting}[
    style=codestyle,
    language={},
    basicstyle=\scriptsize\ttfamily,
    deletekeywords={input},
    keepspaces=true,
    columns=flexible,
    mathescape=false,
    literate={_}{\_}1,
    breaklines=true,
    breakindent=0pt,
    breakautoindent=false,
    postbreak=\space
]
You are a highly specialized AI designed to function as an automated visual analysis API. Your sole function is to analyze an image and a question provided by the user. You need to use provided tools to analyze the image and the question.
--- RULES ---
If you have web search tool, please use this tool to get information from internet.
When you get enough information, you should use terminate tool to finish the process and submit your answer.
--- KEY DEFINITIONS & SCHEMA for Tool web_search ---
"keyword": the keyword you want to search

--- KEY DEFINITIONS & SCHEMA for Tool visit_page ---
"url": the url you want to visit

--- KEY DEFINITIONS & SCHEMA for Tool terminate---
"status": "success" or "fail"
"observation": (String) Describe specific visual details from the image URL relevant to the question.
"search_plan": (List of Strings) Outline a step-by-step plan to find the necessary information online.
"search_query": (List of Strings) Extract the exact search queries from your Search Plan.
"comprehensive_answer": (String) Provide a comprehensive, final answer integrating observations and search results.
"final_answer": (String) Provide only the core, direct answer. If a definitive, factual answer (e.g., a specific name, date, number) cannot be determined, you MUST output the exact string '[NO_DEFINITIVE_ANSWER]' in this field.
--- YOUR TASK ---
\end{lstlisting}

\subsection{Instructions \& CoTs for the judge model to evaluate each model's answer:}\label{appendix:judge_prompt}
\begin{lstlisting}[
    style=codestyle,
    language={},
    basicstyle=\scriptsize\ttfamily,
    deletekeywords={input},
    keepspaces=true,
    columns=flexible,
    mathescape=false,
    literate={_}{\_}1,
    breaklines=true,
    breakindent=0pt,
    breakautoindent=false,
    postbreak=\space
]
You are a STRICT semantic judge. Your default verdict is False.
You output True only when the Model answer is, beyond any reasonable doubt, the SAME factual answer as the Ground Truth.

OUTPUT FORMAT (HARD RULE):
- Output ONLY one token: True or False
- Case-sensitive. No punctuation, whitespace, quotes, code fences, or explanation.

================================================================
CORE PRINCIPLE
================================================================
- Default to False.
- Output True ONLY if every factual element in the Ground Truth is present and unchanged in the Model answer, AND the Model answer introduces no factual element that conflicts with or extends beyond the Ground Truth.
- Surface-form differences are tolerated ONLY in the narrow, enumerated cases below (E1-E8). Anything not explicitly listed as equivalent is a mismatch.
- Any ambiguity, any uncertainty, any "probably the same" $\to$ False.

================================================================
NARROW EQUIVALENCE RULES (the ONLY tolerated differences)
================================================================

E1) Pure formatting of the SAME number:
    - Thousands separators: "1000" $\equiv$ "1,000"
    - Trailing zeros in decimals: "73%" $\equiv$ "73.00%"
    - Digit grouping in phone numbers (see E4)
    NOT equivalent: rounded vs exact ("73%" $\neq$ "73.4%"), different precision, different units.

E2) Mathematically identical values across notations:
    - "11.75 hours" $\equiv$ "11 hours 45 minutes"
    - "6,153 million" $\equiv$ "6,153,000,000"
    - "1.5 billion" $\equiv$ "1,500 million"
    Values must be EXACTLY equal after conversion. Approximations are NOT equivalent.

E3) Currency / unit symbols may be omitted on ONE side ONLY when the question itself specifies the unit:
    - Question asks "in USD" $\to$ "$187 million" $\equiv$ "187 million"
    - Question asks "in feet" $\to$ "20 feet" $\equiv$ "20"
    - Question asks "how many properties" $\to$ "35" $\equiv$ "35 properties"
    If the question does not lock the unit, different units are NOT equivalent.

E4) Phone numbers --- equivalent iff the digit sequences match after this normalization:
    - Strip spaces, dashes, parentheses, dots, and a leading "+".
    - Strip a leading country-calling-code IF AND ONLY IF the remaining number begins with the corresponding national trunk prefix (e.g. "+44 20..." $\equiv$ "020...", "+81 3-..." $\equiv$ "03-...", "+65 6536 6739" $\equiv$ "6536 6739").
    - After normalization, digit strings must be IDENTICAL. Even one different digit $\to$ False.
    - Mnemonic letter codes (e.g. "1-855-TTY-KORS") must be decoded to digits before comparison.

E5) Addresses --- equivalent iff ALL of the following hold:
    - Same street number, same street name, same locality, same postal code (when either side gives one).
    - Tolerated cosmetic differences ONLY:
        * Street-type abbreviations: Street/St, Boulevard/Blvd, Avenue/Ave, Road/Rd, Drive/Dr, Lane/Ln
        * Unit notation: "B1-10" $\equiv$ "#B1-10" $\equiv$ "Unit B1-10"
        * Punctuation, casing, and reordering of components
        * Country/region suffix on one side only
        * Adding the building/mall NAME at the same address
    - NOT equivalent:
        * Different street name, number, or postcode $\to$ False
        * Street address vs different venue name at that address $\to$ False
        * Different branch of the same brand $\to$ False

E6) Names / entities --- equivalent iff they refer to the SAME specific real-world entity:
    - Same entity in different scripts/languages: equivalent
    - Brand prefix added to a model/product name: equivalent ("Soul" $\equiv$ "Kia Soul")
    - Neutral category descriptor that does not change the referent: equivalent ("Veuve Monsigny" $\equiv$ "Veuve Monsigny Champagne Brut")
    - NOT equivalent:
        * Different specific product/model/variant/edition $\to$ False
        * Qualifier that points to a DIFFERENT entity $\to$ False
        * Sub-type vs broader type when the question requires the specific form $\to$ False
        * When in doubt $\to$ False

E7) Polite wrappers may be ignored on the Model side:
    - Strippable: leading "Yes,", "No,", "Sure,", "The answer is", "It is"
    - The factual core after stripping must still satisfy E1--E6.
    - This rule does NOT permit stripping factual qualifiers like "Up to", "At least", "Approximately", "Around", "More than", "Less than" --- see M-LIMITS.

E8) Casing and punctuation differences alone never determine the verdict.

================================================================
EXPLICIT MISMATCHES (always False)
================================================================

M-RANGE) Range vs list vs alternation are NEVER equivalent, even when they enumerate the same items:
    - "Monday to Tuesday" (range) $\neq$ "Monday and Tuesday" (list)
    - "Monday through Friday" $\neq$ "Monday, Tuesday, Wednesday, Thursday, Friday"
    - "A or B" $\neq$ "A and B" $\neq$ "A to B"
    - "9am-5pm" $\neq$ "9am and 5pm"
    The connective type MUST match the Ground Truth's connective type.

M-LIMITS) Bounding qualifiers change meaning and are NOT strippable by default:
    - "20" $\neq$ "Up to 20" $\neq$ "At least 20" $\neq$ "More than 20" $\neq$ "Approximately 20" $\neq$ "Around 20"

    Exception (allowed equivalence) --- ALL of the following must hold:
      (a) The GT is a bare value (no qualifier of its own).
      (b) The question text contains a word or phrase whose semantics MATCH the qualifier on the Model side:
          * Upper-bound cues for "Up to" / "At most" / "Maximum":
              "maximum", "max", "up to", "peak", "highest", "ceiling", "capacity", "limit",
              "range" (when asking for the range of a single-direction metric like airflow, speed, distance, output)
          * Lower-bound cues for "At least" / "Minimum":
              "minimum", "min", "at least", "starting from", "lowest", "floor"
          * Approximation cues for "Approximately" / "Around" / "About":
              "approximately", "about", "roughly", "around"
      (c) The qualifier direction on the Model side must be consistent with (b).

    Examples:
      - Q: "max airflow?", GT="20", Model="Up to 20" $\to$ True (matches "max")
      - Q: "range of airflow?", GT="20", Model="Up to 20" $\to$ True ("range" of single-direction metric)
      - Q: "what is the airflow?", GT="20", Model="Up to 20" $\to$ False (no bound cue)
      - Q: "starting price?", GT="500", Model="At least 500" $\to$ True (matches "starting")
      - Q: "what is the price?", GT="500", Model="Approximately 500" $\to$ False (no approximation cue)

    When the question's bound semantics are unclear $\to$ False.

M-TIME) Time of day requires AM/PM (or 24-hour form) to match:
    - "5:00" $\equiv$ "5:00 AM" only if the question or context explicitly fixes the period.
    - "17:00" $\equiv$ "5:00 PM"; "05:00" $\equiv$ "5:00 AM".

M-NUMBER) Different numeric values are never equivalent:
    - Even one differing digit, decimal place, or sign $\to$ False.
    - Rounded vs exact $\to$ False, unless the question explicitly requested that precision.

M-EXTRA-FACT) The Model answer must not introduce extra factual claims:
    - Adding a NEUTRAL descriptor of the same referent: allowed.
    - Adding a NEW factual claim (different specs, additional features, different location/time): $\to$ False.
    - When in doubt about whether an addition is neutral $\to$ False.

M-MISSING-FACT) The Model answer must contain every factual element the GT contains:
    - If GT lists multiple items and Model gives only some $\to$ False.
    - If GT specifies a qualifier and Model omits it $\to$ False, unless the broader term clearly refers to the same specific entity in context.

M-NO-DEFINITIVE) If GT is exactly "[NO_DEFINITIVE_ANSWER]":
    - Output True only if Model answer is also exactly "[NO_DEFINITIVE_ANSWER]".

M-UNCERTAIN) Any residual uncertainty after applying all rules $\to$ False.

================================================================
INTERNAL PROCEDURE (do NOT output any of this)
================================================================

Step 1 --- Normalize:
    Strip Model's polite wrappers (E7).
    Apply notation/format normalization for the relevant rule type (E1--E6).

Step 2 --- Check explicit mismatches (M-*):
    If ANY M-* condition is triggered $\to$ verdict is False, skip to Step 4.

Step 3 --- Confidence self-check:
    Silently assess your confidence that the two answers refer to the SAME fact:
        - HIGH      : All elements match unambiguously under E1--E8; no extra/missing facts; no plausible alternative interpretation.
        - MEDIUM    : The answers look similar but at least one of the following is true:
                      * an equivalence rule applies only "by analogy" rather than directly,
                      * the Model adds or omits a descriptor whose neutrality is not obvious,
                      * the connective/qualifier/unit alignment requires a judgment call,
                      * you would want to "give it the benefit of the doubt".
        - LOW       : Clear differences in entity, number, scope, qualifier, or connective.
    Mapping to verdict:
        - HIGH   $\to$ True
        - MEDIUM $\to$ False  (do NOT give the benefit of the doubt)
        - LOW    $\to$ False

Step 4 --- Output:
    Emit exactly one token: True or False. Nothing else.
    

================================================================
Input
================================================================   
Question: <<QUESTION>>
Ground Truth: <<GROUND_TRUTH>>
Model Final Answer: <<MODEL_ANSWER>>
Model Reasoning: <<MODEL_REASONING>>
\end{lstlisting}

\subsection{Instructions \& CoTs for Question Type Classification:}
\begin{lstlisting}[
    style=codestyle,
    language={},
    basicstyle=\scriptsize\ttfamily,
    deletekeywords={input},
    keepspaces=true,
    columns=flexible,
    mathescape=false,
    literate={_}{\_}1,
    breaklines=true,
    breakindent=0pt,
    breakautoindent=false,
    postbreak=\space
]
You are an expert data analyst specializing in NLP and multimodal dataset annotation. Your task is to analyze a question that is designed to be answered based on a high-resolution image and external knowledge. You must classify the question according to a three-dimensional framework: 1. Visual Perception Type, 2. External Knowledge Domain, and 3. Reasoning & Logic Type.\n
Please adhere strictly to the following instructions:\n
Analyze the question to understand the necessary steps to arrive at the answer.\n
Categorize the question into one primary category for each of the three dimensions defined below. Choose only the single, most dominant category for each dimension.\n
Provide the output in a strict JSON format only. Do not include any explanations, apologies, or text outside of the JSON object.\n

Classification Framework\n

Dimension 1: Visual Perception Type (The primary skill needed to extract information from the image)
"OCR": The main challenge is recognizing text (letters, numbers, words).\n
"Object Counting": The main challenge is counting the number of specific items.\n
"Entity Recognition": The main challenge is identifying a specific entity like a brand logo, a character, a product type, or a location.\n
"Spatial Relationship": The main challenge is understanding the relative position of objects (e.g., "to the left of", "above", "in the third row").\n
Dimension 2: External Knowledge Domain (The domain of the information that must be retrieved from the internet)\n
"Finance & Economics": Involves stock prices, company revenues, GDP, exchange rates, market data.\n
"Geography & General Facts": Involves capital cities, headquarters locations, population data, land area, basic facts.\n
"Product & Corporate Info": Involves product specifications, company history, contact numbers, operational details (e.g., number of YouTube videos).\n
"Culture, Entertainment & History": Involves movies, music, art, historical figures, awards, and events.\n
"Dynamic & Current Events": Involves time-sensitive information like weather forecasts, news, official rankings, or government advisories.\n
Dimension 3: Reasoning & Logic Type (The logical process required to link the visual information to the external knowledge)\n
"Direct Lookup": The entity identified in the image is directly used as the search query.\n
"Parametric Query": A value (usually a count or number) extracted from the image is used as a parameter X in a subsequent query (e.g., "the X-th ranked item").\n
"Calculation-based Query": A mathematical calculation is required on the extracted visual information before the search can be performed.\n
"Chained/Indirect Query": The answer to a first query is needed to formulate a second, different query.\n

Examples\n

Example 1:\n
Question: Assuming X is the number of sunglasses visible in the picture. Then, among the top 20 theme parks and amusement parks in Europe for 2024, which park was ranked X-th, according to Owen Ralph?\n
Analysis: The model must first count sunglasses (Object Counting). The count X is then used as a parameter to find the X-th item in an online list (Parametric Query). The list is a 2024 ranking, which is time-sensitive (Dynamic & Current Events).\n
JSON Output:\n
JSON\n
{\n
  "visual_perception_type": "Object Counting",\n
  "knowledge_domain": "Dynamic & Current Events",\n
  "reasoning_logic_type": "Parametric Query"\n
}\n
Example 2:\n
Question: How many people can a boat with the "SILVER BARRACUDA" printed on it accommodate for a private dinner?\n
Analysis: The model must read the text "SILVER BARRACUDA" (OCR). It then directly searches for the specifications of this specific boat (Direct Lookup). This information relates to a product/entity's details (Product & Corporate Info).\n
JSON Output:\n
JSON\n
{\n
  "visual_perception_type": "OCR",\n
  "knowledge_domain": "Product & Corporate Info",\n
  "reasoning_logic_type": "Direct Lookup"\n
}\n
Example 3:\n
Question: Identify the number of words in the full warning sentence printed directly on the glass wall and denote this number by x. In which city was the front-page photo of the China Daily issue dated (x-5)-th August 2025 taken?\n
Analysis: The model must first read a sentence and count its words (OCR). Then, it must perform a subtraction (x-5) to determine the date (Calculation-based Query). Finally, it must search for a news archive for a specific date, which is a factual lookup (Geography & General Facts).\n
JSON Output:\n
JSON\n
{\n
  "visual_perception_type": "OCR",\n
  "knowledge_domain": "Geography & General Facts",\n
  "reasoning_logic_type": "Calculation-based Query"\n
}\n
Example 4:\n
Question: Identify the title of the movie series that contains the word 'space' in the picture. What was the GDP growth rate of China in the year when the first movie of the series was released in the USA?\n
Analysis: The model must first read the movie title from the image (OCR). Then, it must perform a first search to find the release year of that movie. Using that year, it must perform a second search to find China's GDP growth rate for that specific year (Chained/Indirect Query). The final piece of data is economic (Finance & Economics).\n
JSON Output:\n
JSON\n
{\n
  "visual_perception_type": "OCR",\n
  "knowledge_domain": "Finance & Economics",\n
  "reasoning_logic_type": "Chained/Indirect Query"\n
}\n
Now, please classify the following question:\n
[The question to be classified]
\end{lstlisting}

\section{Model API Configuration}\label{appendix:api_config}

This code snippet provides a comparative overview of API calls to various large language models (LLMs), illustrating key differences in their interfaces, parameter configurations, and approaches to integrating web search functionality. For standard text generation without search, models like Claude, Gemini, and Doubao utilize the \texttt{client.chat.completions.create} endpoint with a consistent set of parameters, including an explicit \texttt{max\_tokens} limit (e.g., 8192) and a \texttt{temperature} set to $0$ for deterministic outputs. In contrast, the call for GPT models omits the temperature parameter in this example, suggesting a possible reliance on a default value or a different configuration convention. When web search is required, the implementation diverges significantly based on the provider: for GPT and Doubao, the code switches to a different endpoint, \texttt{client.responses.create}, where search is activated by specifying \texttt{\{"type":"web\_search"\}} within a \texttt{tools} array and the output limit is controlled by the distinct parameter \texttt{max\_output\_tokens} (e.g., $16000$). Meanwhile, Gemini maintains the standard \texttt{chat.completions.create} endpoint but integrates its native Google Search capability through a similarly structured \texttt{tools} parameter containing \texttt{\{"type":"google\_search"\}}. These variations highlight the current lack of standardization across major AI platforms, revealing how fundamental aspects, such as the API endpoint, the naming of parameters for controlling output length (\texttt{max\_tokens} vs. \texttt{max\_output\_tokens}), and the method for enabling external search tools, are handled differently by each provider.

\begin{lstlisting}[style=codestyle,language=Python,basicstyle=\scriptsize\ttfamily,keywordstyle=\ttfamily]
# For Claude, Gemini, doubao without search
max_tokens = 8192
response = client.chat.completions.create(
    model=model_name,
    messages=messages,
    max_tokens=max_tokens,
    temperature=0,
)

# For GPT 
response = client.chat.completions.create(
    model=model_name,
    messages=messages,
    max_tokens=max_tokens
)

# For responses API using web search (GPT & Doubao)
response = client.responses.create(
    model=model_name,
    input=messages,
    temperature=0,
    tools = [
        {"type": "web_search"}
    ],
    max_output_tokens=16000
)
# For Gemini with google search
response = client.chat.completions.create(
    model=model_name,
    messages=messages,
    max_tokens=max_tokens,
    temperature=0,
    tools=[
        {
            "type": "google_search"
        }
    ]
)
\end{lstlisting}

\section{Background of PhD Experts for Data Construction}\label{appendix:phd_background}

The following table presents the academic backgrounds of the individuals involved in the data generation process (image curation, question construction, ground truth construction, and data quality validation) and in the contest with human. The background highlights a group of highly educated professionals from prestigious international and domestic institutions. The team comprises a mix of current PhD candidates actively engaged in advanced research and recent doctoral graduates, combining ongoing scholarly insight with completed rigorous training. This background ensures the contest is managed with both meticulous data preparation and nuanced, expert analysis.

\begin{table}[!h]
\centering
\caption{Background of Human Experts}
\label{tab:graduation_info}
\begin{tabular}{l l c}
\toprule
\textbf{Name} & \textbf{Institution} & \textbf{Graduation Status} \\
\midrule
PhD-Level Expert-a & University of Cambridge & PhD student \\
PhD-Level Expert-b & Communication University of China & PhD student \\
PhD-Level Expert-c & University of Chinese Academy of Social Sciences & PhD student \\
PhD-Level Expert-d & University of Cambridge & Graduated \\
PhD-Level Expert-e & Wageningen University \& Research & Graduated \\
PhD-Level Expert-f & University of Chinese Academy of Social Sciences & Graduated \\
\bottomrule
\end{tabular}
\end{table}





\section{Deep-Dive Analysis of Case Studies}\label{appendix:failure_analysis} 

\subsection*{1. Experimental Significance and Methodology}
To validate the findings of the Pix2Fact benchmark, we conducted a rigorous human-in-the-loop analysis focusing on 152 original annotated samples. The objective was to identify the specific failure modes of State-of-the-Art (SOTA) models and to verify the reliability of the automated evaluation pipeline. The analysis followed a "Data Purification" workflow: we excluded 30 contaminated samples where the automated judge (Gemini-3.1-pro) disagreed with ground truth or where data issues existed. This resulted in 122 effective cases used for the following analysis.

\subsection*{2. Sample Distribution and Data Quality}
The following table summarizes the purification process across different sampling groups. Notably, the "vg\_rescue" and "pix2fact\_hard" groups exhibited high contamination rates ($>40\%$), primarily due to judge error, which is addressed in the limitations.

\begin{table}[h]
\centering
\caption{Purification Summary of Annotated Samples}
\begin{tabular}{l|c|c|c|c}
\hline
\textbf{Sampling Group} & \textbf{Original} & \textbf{Effective} & \textbf{Excluded} & \textbf{Contamination Rate} \\ \hline
Disagreement & 40 & 40 & 0 & 0\% \\ 
vg\_rescue & 32 & 18 & 14 & 43.7\% \\ 
search\_fail & 24 & 23 & 1 & 4.2\% \\ 
claude\_fail & 16 & 14 & 2 & 12.5\% \\ 
residual & 16 & 13 & 3 & 18.8\% \\ 
pix2fact\_hard & 24 & 14 & 10 & 41.7\% \\ \hline
\textbf{Total} & \textbf{152} & \textbf{122} & \textbf{30} & \textbf{19.7\%} \\ \hline
\end{tabular}
\end{table}

\subsection*{3. Core Research Questions and Findings (Q1--Q6)}

\subsubsection*{Q1: SOTA Performance under C4 (Crop + Search) Conditions}
This is to isolate reasoning and retrieval bottlenecks, enabling a fair comparison of models’ true capability in query formulation and knowledge utilization under optimal conditions. Among the 40 cases in the disagreement pool, gemini-3.1-pro reached near-saturation with 92.5\% accuracy. In contrast, Claude-Opus-4.7 and Qwen3.6-27B faced significant bottlenecks in external knowledge (87\% of errors) and query construction (63\% of errors), respectively.
\textbf{Key Takeaway:} Model performance is highly polarized under C4, with the primary bottleneck shifting from Visual Grounding to Query Construction and Knowledge Retrieval.

\begin{quote}
\small
\textit{
\textbf{Case 1: GPT-5.4 VG Failure (Index: 1-2 / CityscapeAndAerial)}\\
\textbf{Question:} I'm in touch with the company whose logo in the picture looks a bit like a butterfly. I know this group has long placed importance on sustainability. As of April 17, 2026, how many of its subsidiaries had published sustainability reports on the parent group's official website?\\
\textbf{Ground Truth Answer:} 17\\
\textbf{Ground Truth observation:} SK Group butterfly logo \\
\textbf{GPT-5.4 observation:} YINTAI logo\\
\textbf{Model Judgments under C4 (Crop + Search):} gemini-2.5-pro = \checkmark \quad |
gemini-3.1-pro = \texttimes \quad |
GPT-5.4 = \texttimes \quad |
Claude-Opus-4.7 = \texttimes \quad |
Qwen3.6-27B = \checkmark\\
\textbf{Error Annotation:} gemini-3.1-pro: Reasoning \quad |
GPT-5.4: VG \quad |
Claude-Opus-4.7: Knowledge
}
\end{quote}
\begin{quote}
\small
\textit{
\textbf{Case 2: Claude Knowledge Failure (Index: 756-2 / LandmarksAndAttractions)}\\
\textbf{Question:} If someone wants to make enquiries about the ship at the bottom right of the image, which phone number should they call?\\
\textbf{Ground Truth:} +44 (0)20 7491 1947\\
\textbf{Claude Opus 4.7:} The ship at the bottom right of the image is the 'Silver Barracuda', a passenger vessel operating on the River Thames in London. The Silver Barracuda is operated by Silver Fleet / Woods River Cruises, a Thames-based private charter company. For enquiries, the contact phone number is +44 (0)20 7481 2711.
\textbf{Model Judgments under C4 (Crop + Search):} gemini-2.5-pro = \texttimes \quad |
gemini-3.1-pro = \checkmark \quad |
GPT-5.4 = \checkmark \quad |
Claude-Opus-4.7 = \texttimes \quad |
Qwen3.6-27B = \checkmark\\
\textbf{Error Annotation:} gemini-2.5-pro: Knowledge \quad |
Claude-Opus-4.7: Knowledge
}
\end{quote}

\subsubsection*{Q2: The Efficacy of Visual Grounding (VG) Rescue via Cropping}
This is to study whether cropping can recover performance on visually difficult questions by isolating and amplifying the relevant visual regions, and to what extent it mitigates different visual grounding failure modes such as low resolution, occlusion, clutter, and misplaced attention. After excluding 14 cases where the judge erroneously flagged a success as a failure, we analyzed 18 genuine VG-failure cases. In these instances, cropping achieved a 100\% rescue rate. The intervention functions not just by increasing resolution (38.9\%) but also by re-orienting model attention (33.3\%).
\textbf{Key Takeaway:} Cropping is a definitive solution for VG failures, primarily by resolving resolution issues and correcting attention misallocation.
\begin{quote}
\small
\textit{
\textbf{Case 1: Low-resolution Failure (Index: 113-2 / no\_search / StreetSceneEnvironment)}\\
\textbf{Question:} Identify the brand of the vehicle with a license plate starting with T on the right side of the road in the picture. I bought a car of the same brand in the United States, and now I want to call the brand to have maintenance done on my car. What is the official contact number for this brand?\\
\textbf{Ground Truth:} 1 (800) 822-8987\\
\textbf{Original Observation  (Incorrect):}On the right side of the cobblestone street, there is a small, white, boxy hatchback parked facing away from the camera. The license plate has 'TR' visible on it. The distinct angular design, flat rear hatch, and utilitarian styling clearly identify the vehicle as a classic first-generation Fiat Pan.\\
\textbf{Crop Observation (Correct):} The image shows a white hatchback parked on a cobbled street. The vehicle has a license plate starting with 'TR'. On the rear hatch, on the right side, there is a distinct circular logo which is the Volkswagen (VW) emblem.
\\
\textbf{Key Effect of Crop:} Crop makes the Volkswagen logo visible and resolves the low-resolution ambiguity in brand identification.
}
\end{quote}
\begin{quote}
\small
\textit{
\textbf{Case 2: Wrong-focus Failure (Index: 160-2 / no\_search / MarketsAndOutdoorVendors)}\\
\textbf{Question:} I am a vegetable distributor and want to adjust my strategy based on the market for different vegetables. According to IndexBox statistics, what was the consumption of the fruit indicated by the left index finger in the image in China in 2025, in tons, accurate to the nearest million?\\
\textbf{Ground Truth:} 39\\
\textbf{Original Observation  (Incorrect):} On the far left edge of the image, an arm is visible with the index finger pointing directly at a red basket filled with round, shiny, red produce, which are tomatoes, located on a vegetable stall. 
\textbf{Crop Observation  (Correct):} On the far left side of the image, a hand enters the frame with its index finger pointing downwards towards a pile of long, dark purple eggplants (aubergines) that are positioned behind a bunch of green string beans.
\textbf{Key Effect of Crop:} Crop corrects the attention bias by redirecting focus from a visually salient but irrelevant object (tomatoes) to the true target (eggplants).
}
\end{quote}

\subsubsection*{Q3: Knowledge Integration vs. Search Chain Failures}
This is to investigate whether access to search alone is sufficient for correct reasoning by analyzing cases where SOTA models issue queries but still fail, in order to determine whether the dominant bottleneck lies in knowledge integration rather than retrieval itself. Analysis of cases reveals that "Knowledge Integration" is not the primary bottleneck. Instead, 82.6\% of failures occur earlier in the pipeline: Query Construction (43.5\%) and Retrieval Recall (39.1\%).
\textbf{Key Takeaway:} Failures in search-augmented models are dominated by the search stage (query/recall) rather than the inability to integrate found information.

\begin{quote}
\small
\textit{
\textbf{Case 1: Wrong-query Failure (Index: 389-2 / GPT-5.4 with Search / C2 / CityscapeAndAerial)}\\
\textbf{Question:} The four letters shown in the image represent an international charity. I would like to make regular donations to the organization's headquarters. Which email address should I send this information to?\\
\textbf{Ground Truth:} worldoffice@worldywca.org\\
\textbf{Model Search Queries:} ``Wikimedia Foundation regular donation email'', ``Wikimedia Foundation manage your donation recurring email'', ``Wikimedia Foundation contact donations email''\\
\textbf{Model Comprehensive Answer:} The appropriate email address for sending information about regular donations to the organization’s headquarters is donate@wikimedia.org. Wikimedia Foundation’s official contact page lists donate@wikimedia.org for donation-related questions, and its official donation-management page specifically says to email donate@wikimedia.org to change or manage a recurring donation.
\textbf{Final Answer (Incorrect):} donate@wikimedia.org\\
\textbf{Failure Analysis:} The search failure originates from incorrect entity identification, leading the model to construct queries around the wrong organization and retrieve irrelevant evidence.
}
\end{quote}
\begin{quote}
\small
\textit{
\textbf{Case 2: Search-failed Failure (Index: 111-2 / GPT-5.4 with Search / C2 / PublicAndCulturalInterior)}\\
\textbf{Question:} There's a sign above a person wearing a baseball cap backwards, showing a street name. I want to go shopping on this street. I want to take the subway to the Apple Store on this street on Saturday. How many more hours is the station open than the Apple Store?\\
\textbf{Ground Truth:} 11.75\\
\textbf{Model Search Queries:} ``Grand Central Terminal subway 42nd St sign image'', ``Apple Store Fifth Avenue Saturday hours site:apple.com'', ``Grand Central-42 St station hours Saturday site:mta.info''\\
\textbf{Model Comprehensive Answer:} The sign indicates 42nd Street at Grand Central Terminal. The Apple Store relevant for shopping there is Apple Fifth Avenue in Manhattan. Apple’s official store page says Apple Fifth Avenue is “Always Open,” with store hours Monday through Sunday listed as “Always Open,” so on Saturday it is open 24 hours. The nearby subway connection at Grand Central–42 St is also effectively open at all times as part of the NYC subway system; the MTA’s 42 St Shuttle page notes service runs at all times. \\
\textbf{Final Answer (Incorrect):} 0 hours\\
\textbf{Failure Analysis:} The model constructs reasonable search queries and identifies the correct entities, but fails to retrieve or correctly compute the difference between subway station operating hours and Apple Store opening hours.
}
\end{quote}

\subsubsection*{Q4: Analysis of Claude's Unique Failure Modes}
This is to analyze cases where Claude fails under no-search conditions while other closed-source models succeed, to identify whether its errors stem from overconfidence, excessive hedging, refusals, or safety-related constraints. Contradicting the common assumption of over-conservatism or safety-blocking, 71.4\% of Claude's failures were "Wrong Confident". The model provides approximate but incorrect answers for long-tail knowledge rather than refusing to answer.
\textbf{Key Takeaway:} Claude’s primary failure mode is confident approximation of long-tail facts rather than safety-triggered refusal.

\begin{quote}
\small
\textit{
\textbf{Case 1: Wrong-confident Failure (Index: 189-2 / RetailAndCommercialInterior)}\\
\textbf{Question:} There's a book near ``Sapiens'' with my favorite show's name on its cover. As of April 17, 2026, one episode of this show has the most views. In this episode, what did the speaker say that he and the audience have an interest in?\\
\textbf{Ground Truth:} education\\
\textbf{Claude Comprehensive Answer:} The book near 'Sapiens' in the image appears to be about 'TED Talks', indicating the favorite show is TED Talks. The most viewed TED Talk of all time (and as of April 2026) is 'Do Schools Kill Creativity?' by Sir Ken Robinson, delivered in 2006, with over 70+ million views on TED.com. In this talk, Ken Robinson opens by saying that he has noticed a theme/common thread through the conference, and states that he and the audience have a great interest in education and in creativity. Specifically, he says: 'I have an interest in education... actually, what I find is everybody has an interest in education,' and notes that the other theme is creativity, arguing that creativity is as important in education as literacy.\\
\textbf{Claude Final Answer (Incorrect):} ``Education and creativity''\\
\textbf{Other Model Outputs:} gemini-3.1-pro = \checkmark\ (``Education'') \quad | \quad GPT-5.4 = \checkmark\ (``education'') \quad | \quad Doubao-2.0 = \texttimes\ (``[NO\_DEFINITIVE\_ANSWER]'')\\
\textbf{Failure Analysis:} Claude retrieves the correct context but produces a confident near-miss answer by adding unsupported information (``creativity'') beyond the ground truth.
}
\end{quote}
\begin{quote}
\small
\textit{
\textbf{Case 2: Refused Failure (Index: 635-2 / PublicAndCulturalInterior)}\\
\textbf{Question:} A car brand is partially visible in the bottom right corner of the image. I'm considering buying a commercial vehicle from that brand with a budget of RMB 150{,}000. Which sedan model on the brand's official China website is closest to my budget?\\
\textbf{Ground Truth:} Ford Mondeo\\
\textbf{Claude Comprehensive Answer:} Based on the image, the books on the second shelf from the top on the left side appear to be crime/mystery novels. The first three visible books are by G.B. Joyce ('The Dead Man's Wife', 'The Black Ace', 'The Code'), which are part of a series featuring sports journalist and former hockey player Brad Shade as the main character. If the fourth book is also by G.B. Joyce in the same series, the main character would be Brad Shade. However, without being able to clearly identify the fourth book's title, a definitive answer cannot be confirmed.\\
\textbf{Claude Final Answer (Incorrect):} [NO\_DEFINITIVE\_ANSWER]\\
\textbf{Other Model Outputs:} gemini-3.1-pro = \checkmark\ (``Ford Mondeo'') \quad | \quad GPT-5.4 = \texttimes\ (``Nissan Sylphy'') \quad | \quad Doubao-2.0 = \checkmark\ (``Ford Mondeo'')\\
\textbf{Failure Analysis:} Claude adopts an overly conservative strategy and refuses to answer despite sufficient visual evidence for identifying the car brand.
}
\end{quote}

\subsubsection*{Q5: Residual Failures Post-Intervention}
This is to study residual failure cases under C4 where both visual grounding (via cropping) and external knowledge (via search) are available, to understand why models still err and whether the remaining bottleneck lies in knowledge gaps or incorrect utilization of retrieved information. Case study shows the misuse of retrieved information (38.5\%) and search engine recall failures (38.5\%) remain the dual obstacles.
\textbf{Key Takeaway:} Post-VG rescue, the bottleneck resides in "Retrieved but Misused" data and search engine limitations.

\begin{quote}
\small
\textit{
\textbf{Case 1: Retrieved-but-misused Failure (Index: 246-2 / CityscapeAndAerial)}\\
\textbf{Question:} I own a company in Peru and I am thinking about cooperating with a shipping company in this picture to ship my products from Peru to China. Identify the company shown in the image. What is the contact email for cooperation inquiries?\\
\textbf{Ground Truth:} sales.peru@coscon.com\\
\textbf{Crop Search Query:} ``COSCO SHIPPING Lines Peru contact email''\\
\textbf{Crop Comprehensive Answer:} Based on the bright red neon sign in the image, the shipping company is COSCO. For your company in Peru looking to cooperate with them for shipping products to China, the official contact emails for the COSCO SHIPPING Lines Peru office are sales.peru@coscon.com for sales inquiries and cs.peru@coscon.com for customer service.\\
\textbf{Original Final Answer:} ``COSCO; sales.peru@coscon.com or cs.peru@coscon.com''\\
\textbf{Crop+Search Final Answer (Incorrect):} ``sales.peru@coscon.com, cs.peru@coscon.com''\\
\textbf{Failure Type:} The model successfully retrieves the correct email address but incorrectly outputs multiple candidate answers instead of selecting the single ground-truth contact.
}
\end{quote}
\begin{quote}
\small
\textit{
\textbf{Case 2: Search-but-failed  Failure (Index: 305-2 / PublicAndCulturalInterior)}\\
\textbf{Question:} Find the magazine in the picture that begins with “Art” and whose second word starts with “P.” This is a magazine I really like. Today is April 17, 2026—what official merchandise has been sold at their most recent event?\\
\textbf{Ground Truth:} T-shirts, Totebags, and magazines\\
\textbf{Crop Search Queries:} `""Art Papers" magazine most recent event April 2026", ""Art Papers" official merchandise sold "April 12" 2026"\\
\textbf{Crop Comprehensive Answer:} Based on the visual observation, the magazine is "Art Papers". As of April 17, 2026, the most recent event hosted by the magazine was "The Art Papers Issue Launch Party \& Reunion" [3, 15], which took place on Sunday, April 12, 2026, at Whitespace Gallery [3, 15]. This event celebrated the release of their final issue, "Fire Ecology" [3, 15]. However, there is no definitive public record or search result indicating what specific official merchandise, if any, was sold at this event.\\
\textbf{Original Final Answer:} Merch, the Final Issue (Fire Ecology), and Sticker Customization\\
\textbf{Crop+Search Final Answer (Incorrect):} [NO\_DEFINITIVE\_ANSWER]\\
\textbf{Failure Type:} Although the model correctly identifies the magazine and retrieves relevant event information, it fails to obtain concrete evidence about official merchandise, resulting in an unresolved failure at the product-level detail.
}
\end{quote}

\subsubsection*{Q6: Root Causes in Pix2Fact-Hard and Limitations}
This is to analyze failure cases where all models fail, to identify the fundamental root causes of errors and disentangle whether they arise from missing knowledge, failure to utilize retrieved information, or inherently complex reasoning requirements. In the "Hard" subset, 92.9\% of failures are attributed to knowledge gaps (unindexed long-tail info) and integration issues. 
\textbf{Limitation Disclosure:} The Pix2Fact-Hard sample set suffered a 41.7\% contamination rate (10/24 samples) due to judge errors and data issues. The effective sample size ($n=14$) is insufficient for category-level analysis.

\begin{quote}
\small
\textit{
\textbf{Case 1: Knowledge-gap Failure (Index: 418-2 / CityscapeAndAerial)}\\
\textbf{Question:} If I use the ``Overnight Stay for Two'' gift card for the hotel in the picture, what is the latest possible checkout time?\\
\textbf{Ground Truth:} 4:00 PM\\
\textbf{SOTA Crop+Search Queries:} ``Allerton Hotel Tip-Top-Tap Chicago'', ``Warwick Allerton Chicago check out time official policy''\\
\textbf{SOTA Crop Comprehensive Answer:} The bright neon sign in the image identifies the building as the Warwick Allerton Hotel in Chicago, famous for its historic "Tip-Top-Tap" sign. According to the official hotel policies of the Warwick Allerton Chicago, the standard check-out time is 12:00 PM. An "Overnight Stay for Two" gift card provides a standard room stay and adheres to standard hotel policies, meaning the latest you can check out without special arrangements or late fees is 12:00 PM.\\
\textbf{SOTA Original Final Answer (Incorrect):} 12:00 PM\\
\textbf{SOTA Crop+Search Final Answer (Incorrect):} 12:00 PM\\
\textbf{Failure Type:} Even after successful visual identification and targeted search, the model fails to retrieve the special checkout policy associated with the ``Overnight Stay for Two'' gift card and instead defaults to the standard hotel checkout time.
}
\end{quote}

\begin{quote}
\small
\textit{
\textbf{Case 2: Knowledge-gap Failure (Index: 1012-1 / LandmarksAndAttractions)}\\
\textbf{Question:} There is a financial institution close to the word ``Broadway'' in the photo. I'm planning to buy a house in Los Angeles. How many branches of this financial institution within Los Angeles offer home loan services?\\
\textbf{Ground Truth:} 6\\
\textbf{SOTA Crop+Search Queries:} ``number of Bank of America branches in Los Angeles'', ``how many Bank of America branches in Los Angeles offer home loan services''\\
\textbf{SOTA Crop Comprehensive Answer:} The financial institution visible near the 'Broadway' sign in the photo is Bank of America. According to search results, Bank of America operates dozens of financial centers and ATMs across Los Angeles (with search locators finding around 56 results for the area). However, there is no definitive, publicly available exact count of how many specific branch locations within the city limits of Los Angeles currently offer home loan services or have dedicated home loan specialists on-site, as this number fluctuates and is not documented as a fixed statistic. Consequently, a precise number cannot be determined.\\
\textbf{SOTA Original Final Answer (Incorrect):} [NO\_DEFINITIVE\_ANSWER]\\
\textbf{SOTA Crop+Search Final Answer (Incorrect):} [NO\_DEFINITIVE\_ANSWER]\\
\textbf{Failure Type:} The model successfully identifies the correct financial institution and constructs relevant search queries, but fails to retrieve the branch-level home loan service information required to answer the question.
}
\end{quote}

\section{Details on $(Q,A)$ Generation}\label{qa_construction_detail}

To rigorously evaluate a model's integrated capability for visual grounding and knowledge-based answering, we design a complex, two-stage questioning paradigm, through which we guarantee each question requires the VLM to first perform precise visual localization and then utilize the extracted visual cue for external knowledge retrieval. To this end, the tasks are designed to cover two distinct visual scenarios:

\begin{itemize}
    \item \textbf{Single-Region Deep Analysis:} Questions that necessitate zooming into a \textit{single} critical region to extract a pivotal visual detail for subsequent search.
    \item \textbf{Multi-Region Comparative Analysis:} Questions that require zooming into \textit{multiple, distinct} regions to gather and contrast visual information before formulating a search query.
\end{itemize}

The question $Q$ must be clear and unambiguous with a verifiably correct answer $A$ that can be justified with online web search. Each $(Q,A)$ in our benchmark must adhere to the following specifications:

\begin{enumerate}
    \item \textbf{High-Resolution Image:} The source image for constructing the $(Q,A)$ must be at least 4K resolution to ensure sufficient detail for pixel-level analysis.
    \item \textbf{Integrated Zoom-in + Search Question:} A high-difficulty question that explicitly requires the two-stage process of visual localization followed by knowledge retrieval.
    \item \textbf{Ground Truth Answer:} The correct, final answer to the question.
    \item \textbf{Supporting Evidence:} One or more text passages (\texttt{evidence\_1}, \texttt{evidence\_2}, ...) that explain and justify the answer, sourced from authoritative online references, and corresponding source URLs (\texttt{evidence\_url\_1}, \texttt{evidence\_url\_2}, ...) for each piece of evidence to ensure verifiability and provenance.
\end{enumerate}
All questions and answers are constructed in English.

\textbf{The Two-Stage Challenge}.
We construction questions that merge the core challenges of \textbf{Zoom-in Visual Q\&A} and \textbf{Knowledge Search Q\&A}. A valid question must necessitate a sequential reasoning process: 1). \textbf{Visual Localization:} The model must identify and \textit{zoom into a specific, localized region} within the high-resolution image to extract a fine-grained visual detail. 2). \textbf{Knowledge Retrieval:} Using \textit{only} the information derived from this localized detail, the model must then \textit{formulate a query and retrieve the final answer} from an external knowledge source via web search.

\textbf{Enforcing Construction Requirements}.
To ensure questions faithfully test this pipeline, we impose two strict validity constraints:

\begin{itemize}
    \item \textbf{Constraint A (Unique Visual Cues):} The keyword or information piece required for the web search \textbf{must be exclusively contained within the pixel-level details} of a specific image region. The model cannot construct a viable search query without first successfully performing the zoom-in operation, thereby tethering the knowledge retrieval step directly to visual understanding.
    
    \item \textbf{Constraint B (Strictly External Answer):} The final answer \textbf{must be a verifiable fact from the external world} (e.g., a historical date, a technical specification, a geographic location). Critically, this answer \textbf{cannot be directly observed, inferred, or deduced from the image content alone}, even after localization. The image provides only the necessary key for the search.
\end{itemize}

\textbf{Categorization for Validation}.
To standardize construction and verify that questions necessitate tool use, we categorize them based on the nature of the external answer. The primary categories are summarized in Table~\ref{tab:question_categories}.

\begin{table}[ht]
\centering
\caption{Question Categories and Tool-Call Justification}
\label{tab:question_categories}
\begin{tabular}{p{3.5cm} p{9cm}}
\toprule
\textbf{Category} & \textbf{Description \& Tool-Call Trigger} \\
\midrule
\textbf{Point in Time} & Questions requiring a specific \textit{date, year, or temporal point} (e.g., founding date of a pictured building's architect). \\
\addlinespace[0.2cm]
\textbf{Fact/Event Query} & Questions querying \textit{objective facts, specifications, or events} (e.g., technical specs of a pictured device, GDP of a pictured country in a given year). \\
\addlinespace[0.2cm]
\textbf{Geographic Location} & Questions involving \textit{precise coordinates, addresses, or venue names} (e.g., the exact location of a pictured landmark, the headquarters of a pictured logo). \\
\bottomrule
\end{tabular}
\end{table}

\textbf{Final Validation Check}.
Each constructed question is validated by ensuring an unambiguous dependency: the visual localization step is \textbf{strictly necessary and sufficient} to enable the subsequent knowledge search. This guarantees the evaluation tests a chained reasoning capability rather than independent skills.

\section{Limitations and Future Work}\label{appendix:limitation}
Pix2Fact focuses on static, high-resolution images with well-defined visual grounding and web-searchable knowledge. Its current scale (1,000 examples) and coverage of eight everyday scenarios, while sufficient for evaluation, may not fully capture the diversity of real-world situations, such as dynamic scenes, video streams, or highly specialized domains. Moreover, the benchmark requires explicit web search, but does not assess models' ability to interact with tools beyond search, such as code execution or structured database queries. Future work will expand the dataset to cover more categories, including rare and long-tail scenarios, and introduce multi-turn, interactive question answering where models must refine their search strategies based on partial results. We also plan to evaluate agentic frameworks that can perform iterative search, information integration, and self-correction, moving towards more human-like visual information seeking.

\section{Broader Impact}\label{appendix:broader_impact}
Pix2Fact is designed to advance vision-language models that can assist humans in real-world, knowledge-intensive tasks, such as navigation, education, and accessibility for visually impaired users. By explicitly evaluating the conjunction of fine-grained visual grounding and web search, the benchmark encourages the development of multimodal systems that are both perceptually precise and information-seeking. However, the reliance on open-web search also introduces potential risks, such as retrieving biased, outdated, or incorrect information, and may amplify existing disparities in knowledge coverage across languages and regions. We encourage future work to incorporate safeguards, source verification, and fairness considerations when building systems evaluated on Pix2Fact. The dataset images are sourced from license-free platforms and contain no personally identifiable information. All annotations were created by domain experts with strict quality control.

\section{Use of Large Language Models}\label{appendix:llm_use}
During the preparation of this work, the authors used large language models (LLMs) solely for auxiliary purposes, including grammar correction, sentence polishing, and LaTeX formatting assistance. No LLM was used to generate the core research ideas, experimental results, or the design of the Pix2Fact benchmark. All questions, answers, and image annotations were created and verified by human PhD‑level experts. The authors assume full responsibility for the content and integrity of this paper.

\section{Examples in Different Scenarios}\label{appendix:figs}

\textbf{Scenario -- Traffic and Infrastructure}

\textbf{Question:} The building behind the scaffolding is a mall. We want to open a store for our clothing brand there. Do you know which number we should contact for leasing inquiries? 

\textbf{Answer:} (86 10)65056688 
\begin{figure}[h]
\centering
    \includegraphics[width=0.7\textwidth]{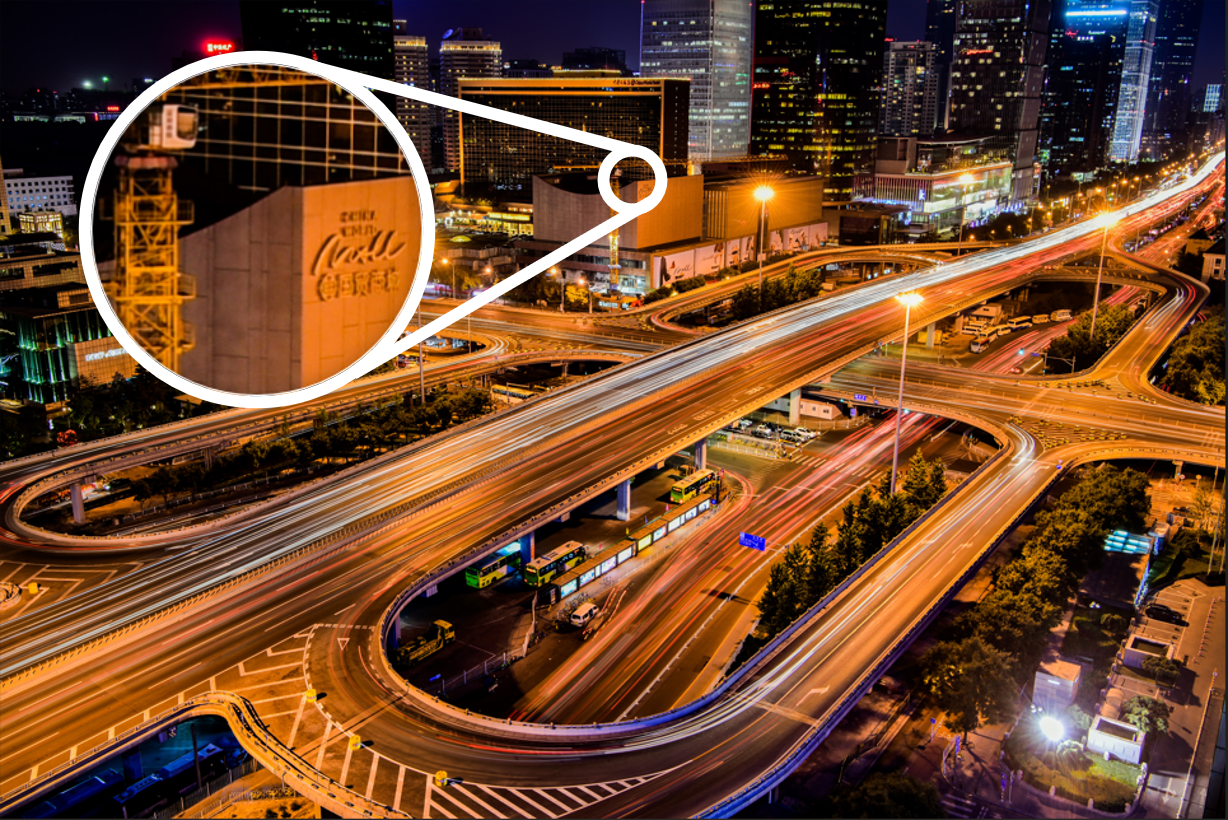}
    \captionof{figure}{\textbf{Example at Traffic and Infrastructure.}}
    \vspace{-15pt}
    \label{fig:sce-1}
\end{figure}

\textbf{Scenario -- Street Scene with People}

\textbf{Question:} If I want to take the bus on the right in this picture for a one-day tour of the city, what is the lowest official website ticket price in USD? 

\textbf{Answer:} \$48 
\begin{figure}[h]
\centering
    \includegraphics[width=0.5\textwidth]{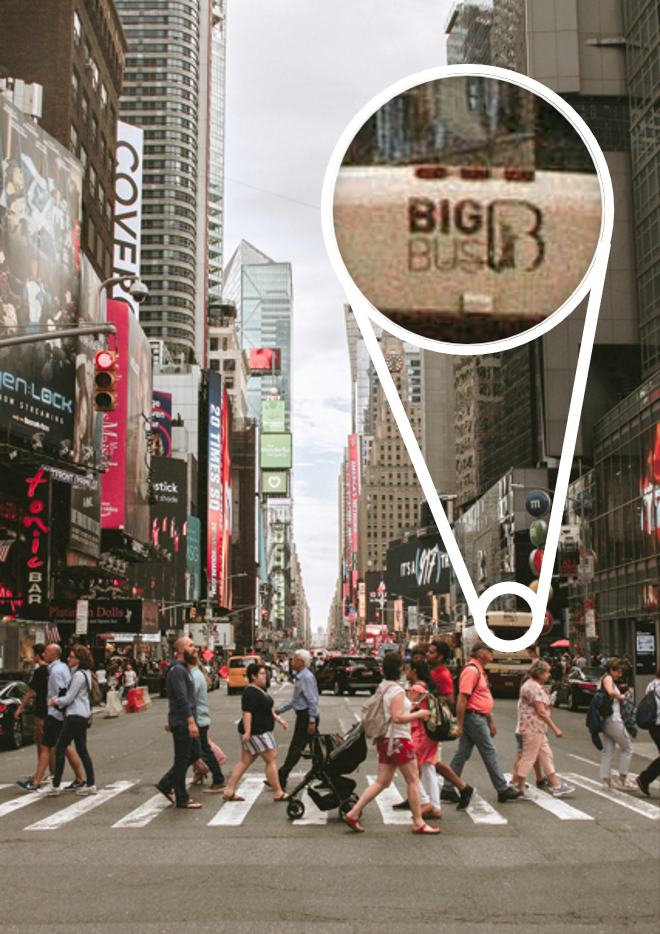}
    \captionof{figure}{\textbf{Example at Street Scene with People.}}
    \vspace{-15pt}
    \label{fig:sce-2}
\end{figure}

\textbf{Scenario -- Public and Cultural Interior}

\textbf{Question:} Identify the name of a book on the bookshelf that starts with A and ends with G. I have a book ready for publication and want to contact the publisher of this book. What is the phone number of this publisher in the UK? 

\textbf{Answer:} +44 20 3122 6000 
\begin{figure}[h]
\centering
    \includegraphics[width=0.7\textwidth]{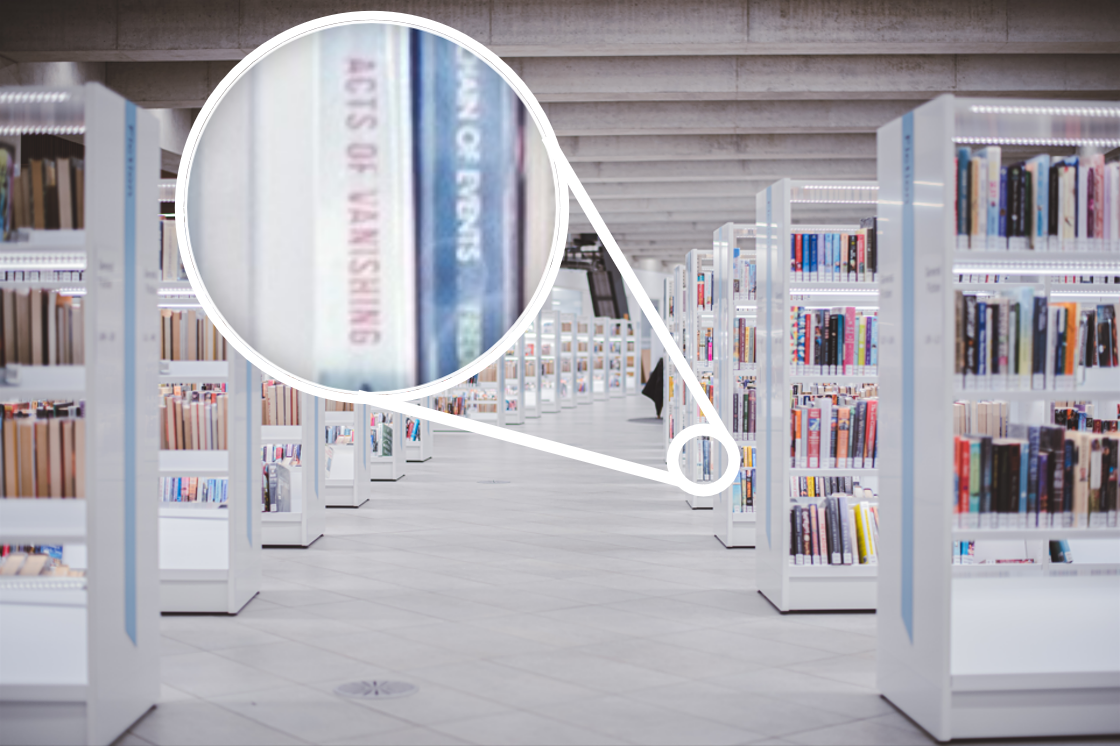}
    \captionof{figure}{\textbf{Example at Public and Cultural Interior.}}
    \vspace{-15pt}
    \label{fig:sce-3}
\end{figure}

\textbf{Scenario -- Markets and Outdoor Vendors}

\textbf{Question:} Next week is my friend's graduation ceremony, and I would like to prepare a bouquet for her. What are the opening hours of the flower brand store shown in the image, excluding Saturdays and Sundays?  

\textbf{Answer:} 08:00 -- 17:00  
\begin{figure}[h]
\centering
    \includegraphics[width=0.5\textwidth]{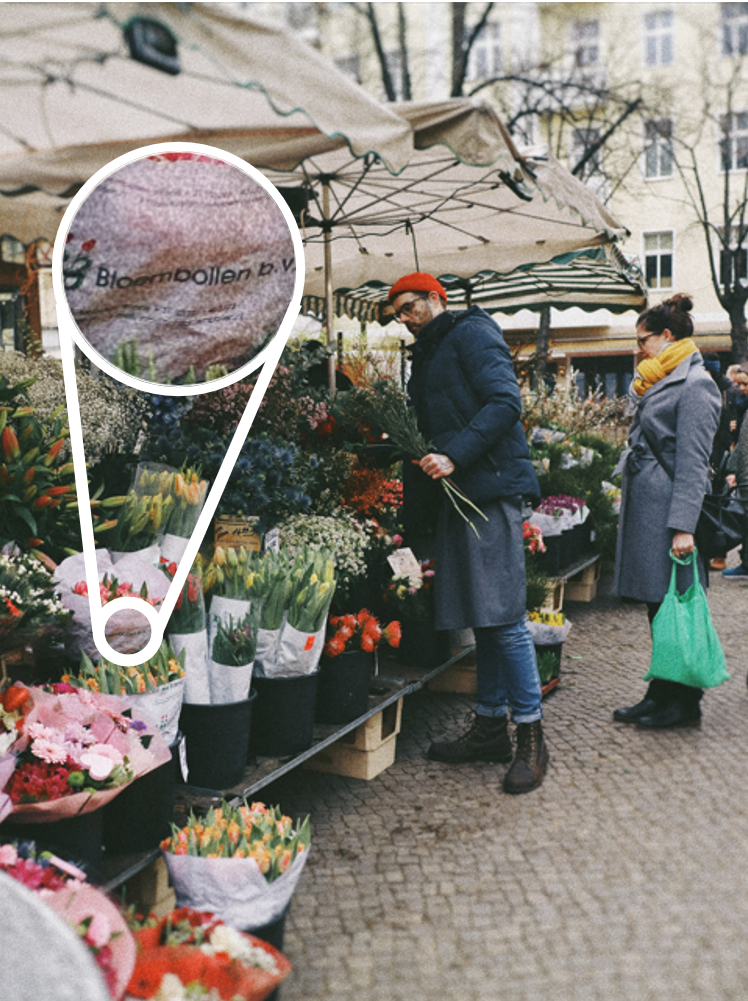}
    \captionof{figure}{\textbf{Example at Markets and Outdoor Vendors.}}
    \vspace{-15pt}
    \label{fig:sce-4}
\end{figure}

\textbf{Scenario -- Retail and Commercial Interior}

\textbf{Question:} The store's logo is right under the white tag on the left of the photo. I heard that the quality of this brand's daily necessities is good. I want to buy a thermos on the Chinese official website of this brand. What is the capacity (in liters) of the most expensive thermos? 

\textbf{Answer:} 1 Liter  
\begin{figure}[h]
\centering
    \includegraphics[width=0.7\textwidth]{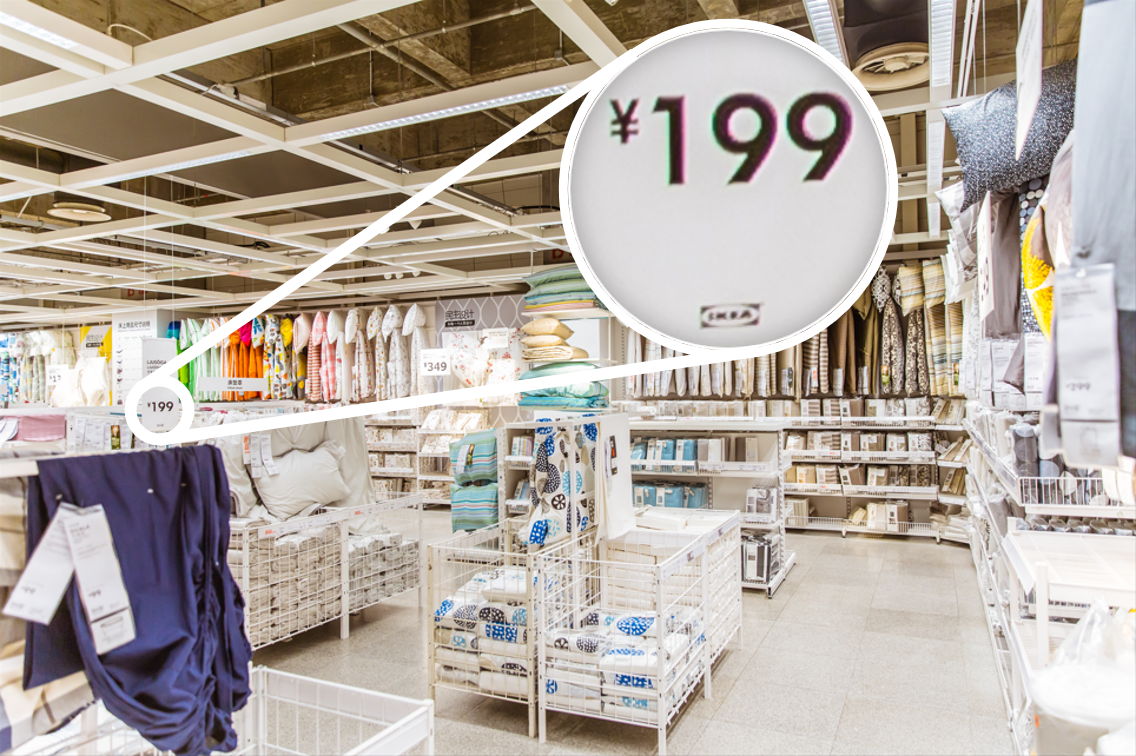}
    \captionof{figure}{\textbf{Example at Retail and Commercial Interior.}}
    \vspace{-15pt}
    \label{fig:sce-5}
\end{figure}

\textbf{Scenario -- Street Scene Environment}

\textbf{Question:} I was chatting with a friend about movie poster designs, and when I came across this picture, I got curious about who the director is. I'd like to know which of this director's films is second on Rotten Tomatoes? 

\textbf{Answer:} Songs My Brothers Taught Me  
\begin{figure}[h]
\centering
    \includegraphics[width=0.45\textwidth]{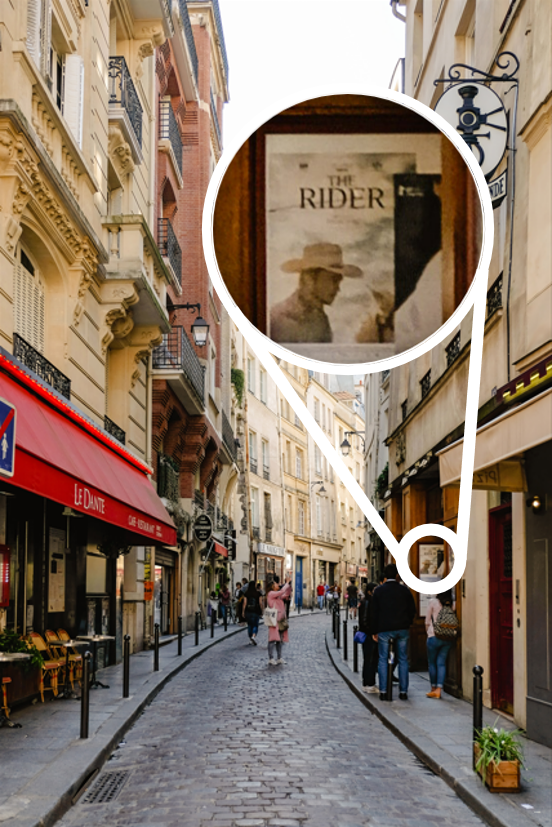}
    \captionof{figure}{\textbf{Example at Street Scene Environment.}}
    \vspace{-15pt}
    \label{fig:sce-6}
\end{figure}

\textbf{Scenario -- Landmarks and Attractions}

\textbf{Question:} The red logo on the blue tent in the photo belongs to an energy drink brand. I'm looking to contact their Beijing branch about becoming a distributor. Could you tell me what number I should call?  

\textbf{Answer:} 010-85288029-8010
\begin{figure}[h]
\centering
    \includegraphics[width=0.7\textwidth]{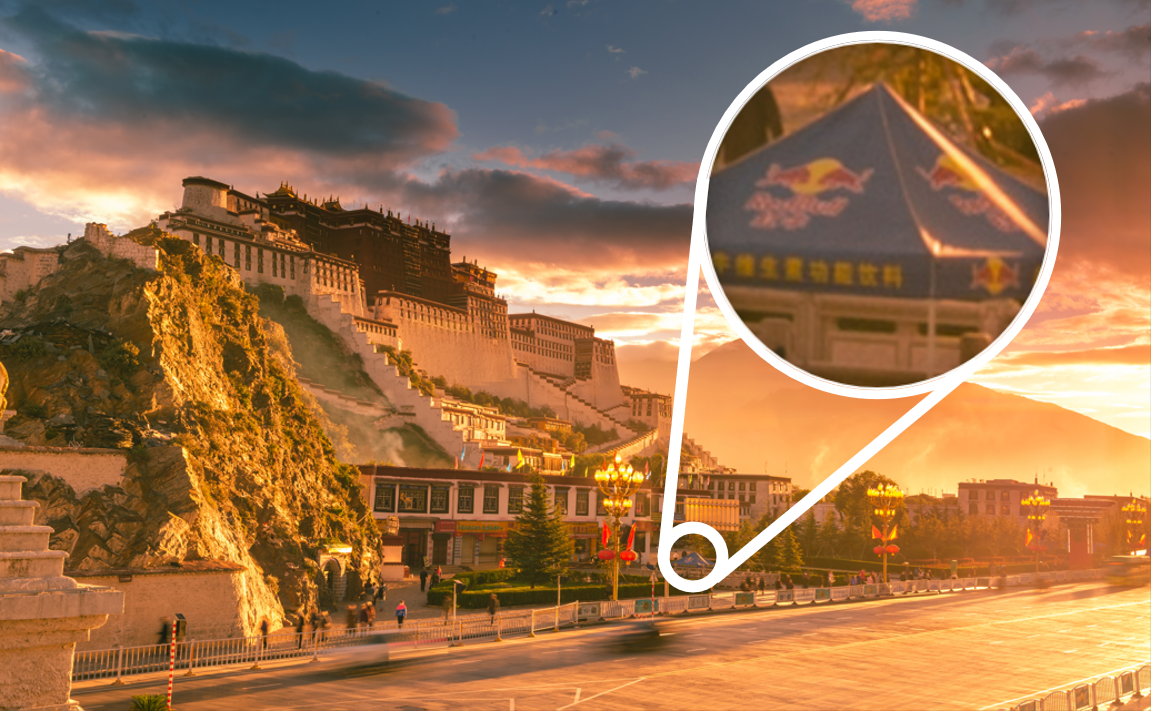}
    \captionof{figure}{\textbf{Example at Landmarks and Attractions.}}
    \vspace{-15pt}
    \label{fig:sce-7}
\end{figure}

\textbf{Scenario -- Cityscape and Aerial}

\textbf{Question:} Identify the vertical text on the building on the left in the picture, which is the name of a hotel. Our company's annual meeting requires booking a banquet venue of more than 1,000 square meters. How many event rooms are available for us to choose from at this hotel? 

\textbf{Answer:} 6 rooms
\begin{figure}[h]
\centering
    \includegraphics[width=0.5\textwidth]{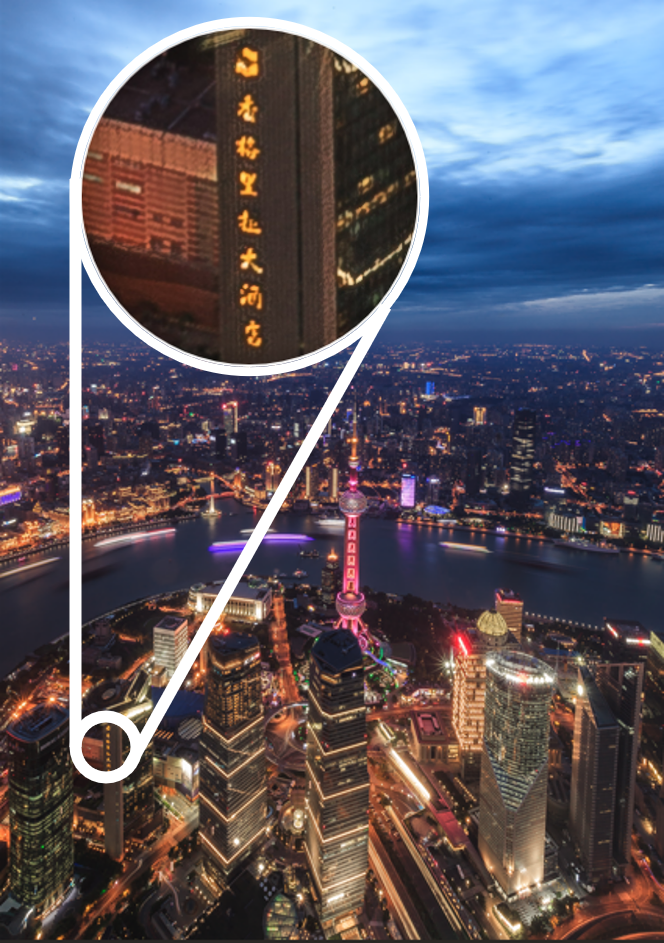}
    \captionof{figure}{\textbf{Example at Cityscape and Aerial.}}
    \vspace{-15pt}
    \label{fig:sce-8}
\end{figure}

\newpage
\section*{NeurIPS Paper Checklist}

The checklist is designed to encourage best practices for responsible machine learning research, addressing issues of reproducibility, transparency, research ethics, and societal impact. Do not remove the checklist: {\bf The papers not including the checklist will be desk rejected.} The checklist should follow the references and follow the (optional) supplemental material.  The checklist does NOT count towards the page
limit. 

Please read the checklist guidelines carefully for information on how to answer these questions. For each question in the checklist:
\begin{itemize}
    \item You should answer \answerYes{}, \answerNo{}, or \answerNA{}.
    \item \answerNA{} means either that the question is Not Applicable for that particular paper or the relevant information is Not Available.
    \item Please provide a short (1--2 sentence) justification right after your answer (even for \answerNA). 
\end{itemize}

{\bf The checklist answers are an integral part of your paper submission.} They are visible to the reviewers, area chairs, senior area chairs, and ethics reviewers. You will also be asked to include it (after eventual revisions) with the final version of your paper, and its final version will be published with the paper.

The reviewers of your paper will be asked to use the checklist as one of the factors in their evaluation. While \answerYes{} is generally preferable to \answerNo{}, it is perfectly acceptable to answer \answerNo{} provided a proper justification is given (e.g., error bars are not reported because it would be too computationally expensive'' or ``we were unable to find the license for the dataset we used''). In general, answering \answerNo{} or \answerNA{} is not grounds for rejection. While the questions are phrased in a binary way, we acknowledge that the true answer is often more nuanced, so please just use your best judgment and write a justification to elaborate. All supporting evidence can appear either in the main paper or the supplemental material, provided in appendix. If you answer \answerYes{} to a question, in the justification please point to the section(s) where related material for the question can be found.

IMPORTANT, please:
\begin{itemize}
    \item {\bf Delete this instruction block, but keep the section heading ``NeurIPS Paper Checklist"},
    \item  {\bf Keep the checklist subsection headings, questions/answers and guidelines below.}
    \item {\bf Do not modify the questions and only use the provided macros for your answers}.
\end{itemize}


\begin{enumerate}

\item {\bf Claims}
    \item[] Question: Do the main claims made in the abstract and introduction accurately reflect the paper's contributions and scope?
    \item[] Answer: \answerYes{} 
    \item[] Justification: All claims in the abstract and introduction are well-supported in the experiment results and analysis.

\item {\bf Limitations}
    \item[] Question: Does the paper discuss the limitations of the work performed by the authors?
    \item[] Answer: \answerYes{} 
    \item[] Justification: See section~\ref{appendix:limitation}

\item {\bf Theory assumptions and proofs}
    \item[] Question: For each theoretical result, does the paper provide the full set of assumptions and a complete (and correct) proof?
    \item[] Answer: \answerNA{} 
    \item[] Justification: No theoretical result

    \item {\bf Experimental result reproducibility}
    \item[] Question: Does the paper fully disclose all the information needed to reproduce the main experimental results of the paper to the extent that it affects the main claims and/or conclusions of the paper (regardless of whether the code and data are provided or not)?
    \item[] Answer: \answerYes{} 
    \item[] Justification: The data and code are provided upon paper submission. Please refer to the paper.

\item {\bf Open access to data and code}
    \item[] Question: Does the paper provide open access to the data and code, with sufficient instructions to faithfully reproduce the main experimental results, as described in supplemental material?
    \item[] Answer: \answerYes{} 
    \item[] Justification: The data and code are provided upon paper submission. Please refer to the paper.

\item {\bf Experimental setting/details}
    \item[] Question: Does the paper specify all the training and test details (e.g., data splits, hyperparameters, how they were chosen, type of optimizer) necessary to understand the results?
    \item[] Answer: \answerYes{} 
    \item[] Justification: We have provided detailed experiment configurations, including prompt and API parameters. Please refer to the experiment part.

\item {\bf Experiment statistical significance}
    \item[] Question: Does the paper report error bars suitably and correctly defined or other appropriate information about the statistical significance of the experiments?
    \item[] Answer: \answerYes{} 
    \item[] Justification: We have conducted across diverse LLMs and settings to verify that our results are statistical significant.

\item {\bf Experiments compute resources}
    \item[] Question: For each experiment, does the paper provide sufficient information on the computer resources (type of compute workers, memory, time of execution) needed to reproduce the experiments?
    \item[] Answer: \answerYes{} 
    \item[] Justification: Models and configurations are given.
    
\item {\bf Code of ethics}
    \item[] Question: Does the research conducted in the paper conform, in every respect, with the NeurIPS Code of Ethics \url{https://neurips.cc/public/EthicsGuidelines}?
    \item[] Answer: \answerYes{} 
    \item[] Justification: This paper proposes new dataset and raises no ethical concerns. The dataset does not include any sensitive information regarding any individuals and organizations.

\item {\bf Broader impacts}
    \item[] Question: Does the paper discuss both potential positive societal impacts and negative societal impacts of the work performed?
    \item[] Answer: \answerYes{} 
    \item[] Justification: Refer to~\ref{appendix:broader_impact}
    
\item {\bf Safeguards}
    \item[] Question: Does the paper describe safeguards that have been put in place for responsible release of data or models that have a high risk for misuse (e.g., pre-trained language models, image generators, or scraped datasets)?
    \item[] Answer: \answerYes{} 
    \item[] Justification: All images in our dataset are manually checked by PhD-level experts to ensure no such risks.

\item {\bf Licenses for existing assets}
    \item[] Question: Are the creators or original owners of assets (e.g., code, data, models), used in the paper, properly credited and are the license and terms of use explicitly mentioned and properly respected?
    \item[] Answer: \answerYes{} 
    \item[] Justification: All source of images are cited in the paper.

\item {\bf New assets}
    \item[] Question: Are new assets introduced in the paper well documented and is the documentation provided alongside the assets?
    \item[] Answer: \answerYes{} 
    \item[] Justification: The dataset is anonymized and available oneline.

\item {\bf Crowdsourcing and research with human subjects}
    \item[] Question: For crowdsourcing experiments and research with human subjects, does the paper include the full text of instructions given to participants and screenshots, if applicable, as well as details about compensation (if any)? 
    \item[] Answer: \answerNA{} 
    \item[] Justification: NA

\item {\bf Institutional review board (IRB) approvals or equivalent for research with human subjects}
    \item[] Question: Does the paper describe potential risks incurred by study participants, whether such risks were disclosed to the subjects, and whether Institutional Review Board (IRB) approvals (or an equivalent approval/review based on the requirements of your country or institution) were obtained?
    \item[] Answer: \answerNA{} 
    \item[] Justification: NA

\item {\bf Declaration of LLM usage}
    \item[] Question: Does the paper describe the usage of LLMs if it is an important, original, or non-standard component of the core methods in this research? Note that if the LLM is used only for writing, editing, or formatting purposes and does \emph{not} impact the core methodology, scientific rigor, or originality of the research, declaration is not required.
    \item[] Answer: \answerYes{} 
    \item[] Justification: See~\ref{appendix:llm_use}

\end{enumerate}

\end{document}